\documentclass{acmart}
\AtBeginDocument{%
  \providecommand\BibTeX{{%
    \normalfont B\kern-0.5em{\scshape i\kern-0.25em b}\kern-0.8em\TeX}}}

\copyrightyear{2022}
\acmYear{2022}
\setcopyright{rightsretained}
\acmConference[FAccT '22]{2022 ACM Conference on Fairness, Accountability, and Transparency}{June 21--24, 2022}{Seoul, Republic of Korea}
\acmBooktitle{2022 ACM Conference on Fairness, Accountability, and Transparency (FAccT '22), June 21--24, 2022, Seoul, Republic of Korea}\acmDOI{10.1145/3531146.3533181}
\acmISBN{978-1-4503-9352-2/22/06}

\usepackage[utf8]{inputenc} 
\usepackage[T1]{fontenc}    
\usepackage{url}            
\usepackage{booktabs}       
\usepackage{amsfonts}       
\usepackage{nicefrac}       
\usepackage{microtype}      
\usepackage{comment}
\usepackage{mathtools} 
\usepackage{setspace}
\usepackage{wrapfig}
\usepackage{lipsum}

\usepackage{amsmath}
\usepackage{natbib}
\usepackage{bm}
\usepackage{xpatch,xcolor}
\usepackage{enumerate}
\usepackage{mathrsfs}
\usepackage{graphicx}
\usepackage{multirow}
\usepackage{tabularx}
\usepackage{array}
\newcolumntype{P}[1]{>{\centering\arraybackslash}p{#1}}
\usepackage{caption}
\usepackage{subcaption}
\usepackage{makecell}

\newtheoremstyle{thm-sf}{}{}{\itshape}{}{\bfseries}{.}{ }{}
\theoremstyle{thm-sf}

\newtheorem{assumption}{Assumption}

\newtheorem{example}{Example}

\newtheorem{proposition}{Proposition}

\DeclareMathOperator{\EX}{\mathbb{E}}

\newcommand{\notepv}[1]{{\color{blue} [{\sffamily\bfseries NOTE PV:} {\em #1}]}}

\newcommand{\newar}[1]{{\color{black} #1}}
\newcommand{\notear}[1]{{\color{magenta} [{\sffamily\bfseries NOTE AR:} {\em #1}]}}

\usepackage{multicol}
\newcommand{\vX}{\bm X} 
\newcommand{\vx}{\bm x} 
\newcommand{\sX}{\mathcal X} 
\newcommand{\sQ}{\mathcal Q} 
\newcommand{\sG}{\mathcal G} 
\renewcommand{\G}{G} 
\newcommand{\g}{g} 
\newcommand{\sR}{\mathcal R} 
\newcommand{\q}{q}  
\renewcommand{\r}{r}  
\newcommand{\R}{R}  
\newcommand{\vM}{\bm M}  
\newcommand{\sF}{\mathcal F}  
\newcommand{\sM}{\mathcal M}  
\newcommand{\f}{F}  
\newcommand{\vF}{\bm F}  
\newcommand{\m}{M}  
\renewcommand{\P}{p}  
\newcommand{\sP}{\mathcal P}  

\begin{document}

\title[Learning Resource Allocation Policies from Observational Data]{Learning Resource Allocation Policies from Observational Data with an Application to Homeless Services Delivery}



\author{Aida Rahmattalabi}
\affiliation{%
  \institution{University of Southern California}
  \country{USA}
  }
\email{rahmatta@usc.edu}

\author{Phebe Vayanos}
\affiliation{%
  \institution{University of Southern California}
  \country{USA}
}
\email{phebe.vayanos@usc.edu}

\author{Kathryn Dullerud}
\affiliation{%
  \institution{University of Southern California}
  \country{USA}
}
\email{kdulleru@usc.edu}

\author{Eric Rice}
\affiliation{%
  \institution{University of Southern California}
  \country{USA}
}
\email{ericr@usc.edu}


\begin{abstract}
We study the problem of learning, from observational data, fair and interpretable policies that effectively match heterogeneous individuals to scarce resources of different types. We model this problem as a multi-class multi-server queuing system where both individuals and resources arrive stochastically over time. Each individual, upon arrival, is assigned to a queue where they wait to be matched to a resource. The resources are assigned in a first come first served (FCFS) fashion according to an \emph{eligibility structure} that encodes the resource types that serve each queue. We propose a methodology based on techniques in modern causal inference to construct the individual queues as well as learn the matching outcomes and provide a mixed-integer optimization (MIO) formulation to optimize the eligibility structure. The MIO problem maximizes policy outcome subject to wait time and fairness constraints. It is very flexible, allowing for additional linear domain constraints. We conduct extensive analyses using synthetic and real-world data. In particular, we evaluate our framework using data from the U.S. Homeless Management Information System (HMIS). We obtain wait times as low as an FCFS policy while improving the rate of exit from homelessness for underserved or vulnerable groups (7\% higher for the Black individuals and 15\% higher for those below 17 years old) and overall.
\end{abstract}




\maketitle



\section{Introduction}

We study the problem of designing policies to effectively match heterogeneous individuals to scarce resources of different types. We consider the case where both individuals and resources arrive stochastically over time. Upon arrival, each individual is assigned to a queue where they wait to be matched to a resource. This problem arises in several public systems such as those providing social services, posing unique challenges at the intersection of {efficiency and fairness. In particular, the joint characteristics of individuals and their matched resources determine the effectiveness of an allocation policy, making it crucial to match individuals with the right type of resource. Furthermore, when a resource becomes available, a decision-maker should decide whom among the individuals waiting in various queues should receive the resource which impacts the wait time of different individuals. In addition, since there are insufficient resources to meet demand, there are inherent fairness considerations for designing such policies.}


We are particularly motivated by the problem of allocating housing resources among individuals experiencing homelessness. According to the U.S. Department of Housing and Urban Development (HUD), more than 580,000 people experience homelessness on a given night~\cite{Henry2020AHAR:Exchange}. The Voices of Youth Count study found youth homelessness has reached a concerning prevalence level in the United States; one in 30 teens (13 to 17) and one in 10 young adults (18 to 25) experience at least one night of homelessness within a 12-month period, amounting to 4.2 million persons a year~\cite{Morton2018PrevalenceStates}. Housing interventions are widely considered as the key solution to address homelessness~\cite{U.S.Dept.ofHousingandUrbanDevelopment2007TheReport}. In the U.S., the government funds programs that assist homeless using different forms of housing interventions and services~\cite{2015OpeningHomelessness}. The HMIS database collects information on the provision of these services.

\begin{figure}[t!]
    \centering
    \includegraphics[width = 0.35\textwidth]{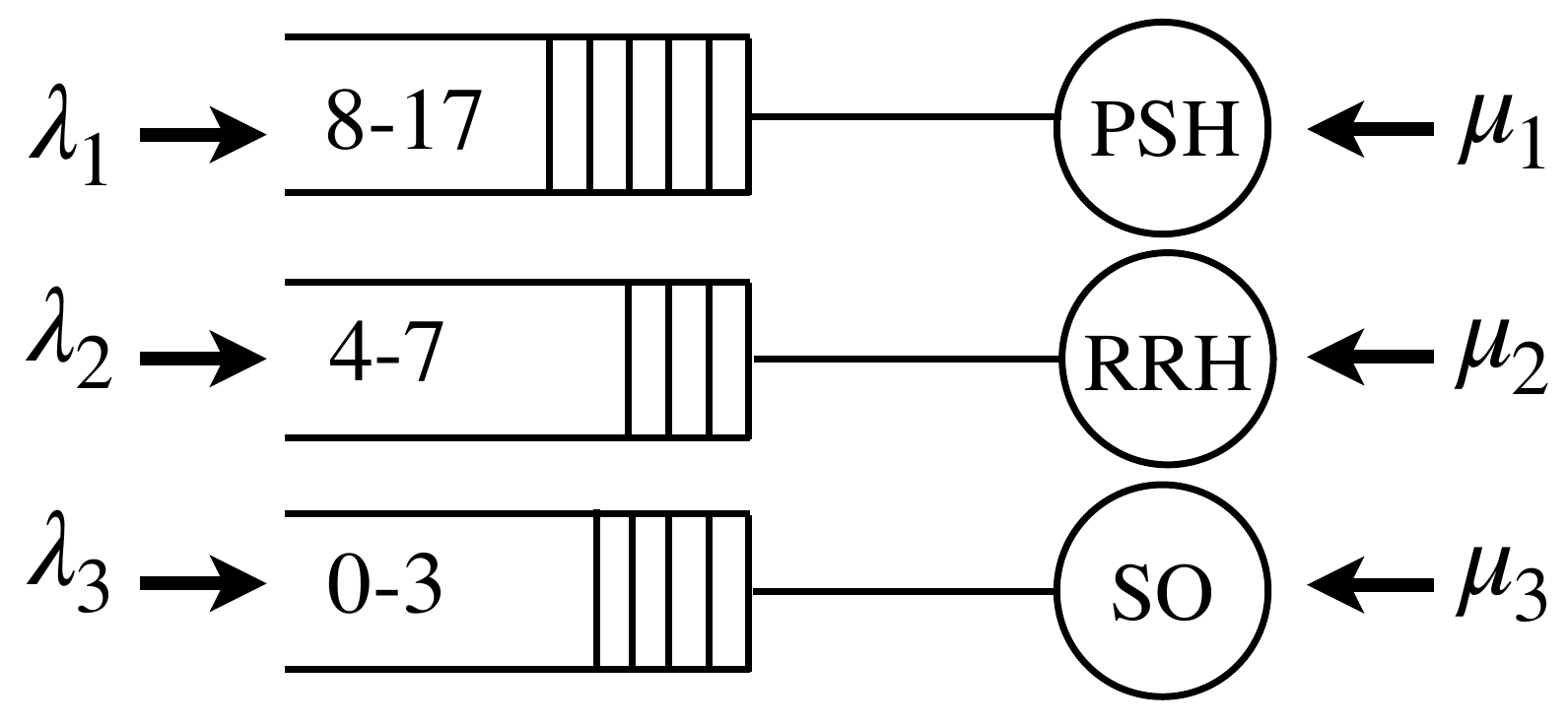}
    \caption{NST-recommended resource allocation policy utilized by housing allocation agencies in the homelessness context.
    The policy is in the form of a resource eligibility structure. According to this figure, individuals with score eight and above qualify for PSH, score 4 to 7 are assigned to the RRH wait list and finally individuals who score below 4 are not assigned to any of the housing interventions.}
    \label{fig:HMIS-matching}
    \vspace*{-\baselineskip}
\end{figure}

Unfortunately, the number of homeless individuals in the U.S. far exceeds the available resources which necessitates strategic allocation to maximize the intervention's effectiveness. Many communities have attempted to address this problem by creating coordinated community responses, typically referred to as Coordinated Entry Systems (CES). In such systems, most agencies within a community pool their housing resources in a centralized system called a Continuum of Care (CoC). A CoC is a regional or local planning body that coordinates housing and services funding—primarily from HUD—for people experiencing homelessness.
Individuals in a given CoC who seek housing are first assessed for eligibility and vulnerability and those identified as having the greatest need are matched to appropriate housing resources~\cite{Rice2017AssessmentYouth}. For example, in the context of youth homelessness, the most widely adopted tool for assessing vulnerability is the Transition Age Youth-Vulnerability Index-Service Prioritization Decision Assistance Tool (TAY-VI-SPDAT): Next Step Tool (NST), which was developed by OrgCode Consulting, Corporation for Supportive Housing (CSH), Community Solutions, and Eric Rice. OrgCode claims that hundreds of CoC's in the USA, Canada and Australia have adopted this tool~\cite{Orgcode2015TransitionYouth}. After assessment, each individual receives a vulnerability score ranging from 0 to 17. One of the main challenges that CoC's face is how to use the information about individuals to decide what housing assistance programs should be available to a particular homeless individual. In many communities, based on the recommendations provided in the NST tool documentation, individuals who score 8 to 17 are considered as ``high risk'' and are prioritized for resource-intensive housing programs or Permanent Supportive Housing (PSH). Those who score in the 4-7 range are typically assigned to short-term rental subsidy programs or Rapid-ReHousing (RRH) and those with score below 4 are eligible for services that meet basic needs which we refer to as Service Only (SO)~\cite{Rice2018LinkingYouth}. Figure~\ref{fig:HMIS-matching} depicts how the individuals are matched to resources according to the status-quo policy.

The aforementioned policy can be viewed as a \emph{resource eligibility structure} as from the onset, it determines the resources an individual is eligible for. Such policies have the advantage of being interpretable, i.e., it is easy to explain why a particular allocation is made. Earlier work shows that most communities follow the policy recommendations when assigning housing~\cite{Rice2018LinkingYouth}. However, controversy has surrounded the use of these cut scores and as of December 2020, OrgCode has called for new approaches to using the data collected by HMIS~\cite{OrgCode2020TheVI-SPDAT}. There is also an overwhelming desire on the part of HUD to design systematic and data-driven housing policies, including the design of the cut scores and the queues that they induce~\cite{2015OpeningHomelessness}. Currently, the cut scores are not tied to the treatment effect of interventions or the relative arrival rate of individuals and resources in the respective queues. This is problematic as it is not evidently clear that assigning high-scoring and mid-scoring individuals to particular housing
interventions, such as PSH or RRH, actually increases their chances of becoming stably housed. Additionally, there may not be enough resources to satisfy the needs of all individuals matched to a particular resource, resulting in long wait times. Prolonged homelessness may in turn increase the chances of exposure to violence, substance use, etc., or individuals dropping out of the system. 

In particular, OrgCode and others have called for a new equity focus to how vulnerability tools are linked to housing allocation \cite{OrgCode2020TheVI-SPDAT,Milburn2021InequityHomelessness}. Despite recent efforts to understand and mitigate disparities in homelessness, current system suffers from a significant gap in the prevalence of homelessness across different groups. 
For example, studies show that most racial minority groups experience homelessness at higher rates than Whites~\cite{Fusaro2018RacialStates}. Also, recent work has revealed that PSH outcomes are worse for Black clients in Los Angeles~\cite{Milburn2021InequityHomelessness} and based on the same HMIS data used in present study, Black, Latinx, and LGBQ youth have been shown to experience worse housing outcomes~\cite{Hill2021AnStatus}. Addressing these disparities requires an understanding of the distribution of the individuals vulnerability to homelessness, the heterogeneity in the treatment affect and the associations with protected attributes such as race, gender, or age.

In this work, we build on the literature on causal inference and queuing theory and propose a methodology that uses historical data about the waitlisted individuals and their allocated resources to optimize resource allocation policies. 
We make the following contributions:
\begin{itemize}
    \item We model the policy optimization problem as a multi-class multi-server queuing system between heterogeneous individuals and resources that arrive over time. We extend the literature on queuing theory by proposing a data-driven methodology to construct the model from observational data. Specifically, we use tools from modern causal inference to learn the treatment effect of the interventions from data and construct the queues by grouping individuals that have similar average treatment effects.  
    \item We propose interpretable policies that take the form of a {resource eligibility structure}, encoding the resource types that serve each queue.
    We provide an MIO formulation to optimize the eligibility structure that incorporates flexibly defined fairness considerations or other linear domain-specific constraints. The MIO maximizes the policy effectiveness 
    and guarantees minimum wait time.
    \item Using HMIS data, we conduct a case study to demonstrate the effectiveness of our approach. Our results indicate superior performance along policy effectiveness, fairness and wait time. Precisely, we are able to obtain wait time as low as a fully FCFS policy while improving the rate of exit from homelessness for traditionally underserved or vulnerable groups (7\% for the Black individuals and 15\% higher for youth below 17 years old) and overall. 
\end{itemize}
The remainder of this paper is organized as follows. In Section~\ref{sec:literature}, we review the related literature. 
In Section~\ref{sec:statement}, we introduce the policy optimization problem. In Section~\ref{sec:solution}, we propose our data-driven methodology for solving the policy optimization problem. Finally, we summarize our numerical experiments and present a case study using HMIS data on youth experiencing homelessness in Section~\ref{sec:experiments}. Proofs and detailed numerical results are provided in the Appendix.



\section{Literature Review}\label{sec:literature}

This work is related to several streams of literature which we review. Specifically, we cover queuing theory as the basis of our modelling framework. We also position our methodology within the literature on data-driven policy optimization and causal inference. We conclude by highlighting recent works on fairness in resource allocation. 

A large number of scarce resource allocation problems give rise to one-sided queuing models. In these models, resources are allocated upon arrival, whereas individuals queue before being matched. Examples are organ matching~\cite{Bandi2019RobustSystems} and public housing assignment~\cite{Kaplan1984ManagingHousing,Arnosti2020DesignAllocation}.
One stream of literature studies dynamic matching policies to find asymptotically optimal scheduling policies under conventional heavy traffic conditions~\cite{Mandelbaum2004SchedulingC-rule, Ata2013OnCosts}. Another stream focuses on the system behavior under FCFS service discipline aiming to identify conditions that ensure the stability of the queuing system and characterize the steady-state matching flow rates, i.e., the average rate of individuals of a given queue (or customer class) that are served by a particular resource (server)~\cite{Fazel-Zarandi2018ApproximatingLaw, Castro2020MatchingSolution}. 
These works only focus on minimizing delay and do not explicitly model the heterogeneous service value among the customers.
Recently,~\cite{Ding2021ARewards} studied one-sided queuing system where resources are allocated to the customer with the highest score (or index), which is the sum of the customer’s waiting score and matching score. The authors derive a closed-form index that optimizes the steady-state performance subject to specific fairness considerations. Their proposed fairness metric measures the variance in the likelihood of getting service before abandoning the queue. Contrary to their model, we consider FCFS policies subject to resource eligibility structures which we optimize over. Our model is based on the policies currently being implemented for housing allocation among homeless individuals \newar{that target resources to heterogeneous individuals by explicitly imposing an eligibility structure}. \newar{Our policies are interpretable by design as upon arrival the resources that an individual is eligible for is known, making it easy to explain why a certain allocation has or has not been made}. Further, the proposed model allows for a more general class of fairness requirements commonly used in practice including fairness in \emph{allocation} and \emph{outcome}. \newar{It is noteworthy that our model is different from common allocation models in the public housing setting where targeting of the resources can be considered as implicit, i.e., individuals with different levels of need make different choices about where to apply and what to accept themselves~\cite{Arnosti2020DesignAllocation, Arnosti2019HowHousing}}.

Our approach builds upon~\cite{Afeche2021OnSystem}, in which the authors study the problem of designing a matching topology between customer classes and servers under a FCFS service discipline. They focus on finding matching topologies that minimize the customers' waiting time and maximize matching rewards obtained by pairing customers and servers. The authors 
characterize the average steady-state wait time across all customer classes in terms of the structure of the matching model, under heavy-traffic condition. 
They propose a quadratic program (QP) to compute the steady-state matching flows between customers and servers and prove the conditions under which the approximation is exact. We build on the theoretical results in~\cite{Afeche2021OnSystem} to design resource eligibility structures that match heterogeneous individuals and resources in the homelessness setting. Contrary to the model  in~\cite{Afeche2021OnSystem}, we do not assume that the queues or the matching rewards are given a priori. Instead, we propose to use observational data from historical policy to learn an appropriate grouping of individuals into distinct queues, estimate the matching rewards, and evaluate the resulting policies.

Another stream of literature focuses on designing data-driven policies, where fairness considerations have also received significant attention due to implicit or explicit biases that models or the data may exhibit~\cite{Bertsimas2013FairnessTransplantation,Dickerson2015FutureMatch:Environments,Rahmattalabi2021FairApproach, Keymanesh2021Fairness-awareDecision-Making}. In~\cite{Bertsimas2013FairnessTransplantation}, the authors propose a data-driven model for learning scoring policies for kidney allocation that matches organs at their time of procurement to
available patients. Their approach satisfies linear fairness constraints approximately and does not provide any guarantees for wait time. In addition, they take as input a model for the matching rewards (i.e., life years from transplant)to optimize the scoring policy.
In~\cite{Azizi2018DesigningResources}, the authors propose a data-driven mixed integer program with linear fairness constraints to solve a similar resource allocation which provides an exact, rather than an approximate, formulation. They also give an approximate solution to achieve faster run-time. We consider a class policies in the form of matching topologies that is different from scoring rules and is more closely related to the policies implemented in practice. Such policies offer more interpretability as individuals know what resources they are eligible for from the onset. Several works have considered interpretable functional forms in policy design. For example, in~\cite{Bertsimas2019OptimalTrees, Jo2021LearningData}, the authors consider decision trees and develop techniques to obtain optimal trees from observational data. Their approach is purely data-driven and do not allow for explicit modelling of the arrival of resources, individuals which impact wait time. In the homelessness setting, our work is closely related to~\cite{Kube2019} which proposes a resource allocation mechanism to match homeless households to resources based on the probability of system re-entry. In this work, the authors provide a static formulation of the problem which requires frequent re-optimization and does not take the waiting time into account. In~\cite{Nguyen2021ScarceJustice}, the authors propose a fairness criterion that prioritizes those who benefit the most from a resource, as opposed to those who are the neediest and study the price of fairness under different fairness definitions. Similar to~\cite{Kube2019}, their formulation is static and does not yield a policy to allocate resources in dynamic environments. 

\vspace{-2mm}

\section{Housing Allocation as a Queuing System}\label{sec:statement}

\subsection{Preliminaries}
We model the resource allocation system as an infinite stream of heterogeneous individuals and resources that arrive over time. Each individual is characterized by a (random) feature vector $\vX \in \sX \subseteq \mathbb R^n$ and receives an intervention $\R$ from a finite set of treatments indexed in the set $\sR$. We note that $\sR$ may include ``no intervention'' or minimal interventions such as SO in the housing allocation setting.
Using the potential outcomes framework~\cite{Rubin2005CausalOutcomes}, each individual has a vector of potential outcomes $Y(r) \in \mathcal Y \subseteq \mathbb R \; \forall \r \in\sR$, where $Y(\r)$ is an individual's outcome when matched to resource $\r$. 


We assume having access to $N$ historical observations $\mathcal D := \left\{(\bm X_i, R_i, Y_i)\right\}^{N}_{i=1}$, generated by the deployed policy, where $\vX_i \in \sX$ denotes the feature vector of the $i$th observation, $\R_i \in \sR$ is the resource assigned to it and $Y_i = Y_i(\R_i)$ is the observed outcome, i.e., the outcome under the resource received. {A (stochastic) policy $\pi(\r|\vx): \sX \times \sR \rightarrow [0, 1]$ maps features $\vx$ to the probability of receiving resource $\r$. We define the value of a policy as the expected outcome when the policy is implemented, i.e.,
$V(\pi) := \EX[\sum_{r \in \mathcal R}\pi(r | \bm X)Y(r)]$.}
A major challenge in evaluating and optimizing policies is that we cannot observe the counterfactual outcomes $Y_i(r), r \in \sR, r \neq \R_i$ of resources that were not received by data point $i$. Hence, we need to make further assumptions to identify policy values from historical data. In Section~\ref{sec:solution}, we elaborate on these assumptions and propose our methodology for evaluating and optimizing policies from data.

We model the system as a multi-class multi-server (MCMS) queuing system where a set of resources $\sR$ serve a finite set of individual queues indexed in the set $\sQ$. Upon arrival, individuals are assigned to different queues based on their feature vector. For example, in the housing allocation setting and according to the recommended policy the assignment is based on the vulnerability score. We use $\P : \sX \rightarrow \sQ$ to denote the function that maps the feature vector to a queue that the individual will join. We refer to $\P$ as the \emph{partitioning function} (as it partitions $\sX$ and assigns each subset to a queue) and note that it is unknown a priori. In this work, we consider partitioning functions in the form of a binary trees similar to classification trees, due to their interpretability~\cite{Azizi2018DesigningResources}. We assume that individuals arrive according to stationary and independent Poisson processes and that inter-arrival time of resources follows an exponential distribution. \newar{These are common assumptions in queuing theory for modeling arrivals, however, they may not fully hold in practice in which case, it is possible to use re-optimization to adapt to the changing environment.}
We use $\bm \lambda := (\lambda_1, \dots, \lambda_{|\sQ|})$ and $\bm \mu := (\mu_1, \dots, \mu_{|\sR|})$ to denote the vector of arrival rates of individuals and resources, respectively. We define $\lambda_{\sQ} := \sum_{\q \in \sQ} \lambda_{\q}$ and $\mu_{\sR} := \sum_{\r \in \sR} \mu_{\r}$ as the cumulative arrival rates of individuals and resources, respectively. Without loss of generality, we assume that  $\lambda_\q > 0 \; \forall \q \in \sQ$ and $\mu_\r > 0 \; \forall \r \in \sR$.

\subsection{Matching Policy}
Once a new resource becomes available, it is allocated according to a resource eligibility structure that determines what queues are served by any particular resource. The resource eligibility structure can be  represented as a matching topology $\vM := [\m_{\q\r}] \in \{0,1\}^{|\sQ| \times |\sR|},$ where $\m_{\q\r} = 1$ indicates that individuals in queue $\q$ is eligible for resource $\r$. Resources are assigned to queues in an FCFS fashion subject to matching topology $\vM$. For a partitioning function $\P$ and matching topology $\vM$, we denote the allocation policy by $\pi_{\P, \vM}(\r | \vx)$. We concern ourselves with the long-term steady state of the system. Proposition~\ref{prop:steady-state} gives the necessary and sufficient conditions to arrive at a steady-state. 

\begin{proposition}[\citet{Adan2014AAbandonments}, Theorem 2.1]
Given the MCMS system defined through $(\mathcal Q, \mathcal R, \bm \lambda, \bm \mu, \vM)$, under the FCFS service discipline matching $\vM$ admits a steady state if and only if the following condition is satisfied:
\begin{equation*}
    \mu_{\mathscr{R}} - \sum_{\r \in \mathscr{R}}\sum_{\q \in \sQ_{\mathscr{R}}(\vM)} \lambda_{\q} > 0 \quad \forall \mathscr{R} \subseteq \sR.
\end{equation*}
The left-hand side is the cumulative arrival rate of resources in $\mathscr{R}$ in excess of the cumulative arrival rate of all the queues in $\sQ_{\mathscr{R}}$, where $\sQ_{\mathscr{R}}$ is the set of queues that are only eligible for resources in $\mathscr{\R}$, i.e., $\sQ_{\mathscr{R}} = \{\q \in \sQ : \sum_{\r \in \sR\setminus\mathscr{R}}\m_{\q\r} = 0\}$.
\label{prop:steady-state}
\end{proposition}
We define the set of \emph{admissible matching topologies} as those that satisfy the inequality in Proposition~\ref{prop:steady-state}. 
In the housing allocation problem, we assume that SO resources are abundant, i.e., $\mu_{\sR} - \lambda_{\sQ} > 0 $. \newar{In other settings, it is possible to create an auxiliary resource queue corresponding to no intervention.} The abundance assumption ensures that there exists at least one admissible matching: the fully connected matching topology $\m_{\q\r} = 1\; \forall \q \in \sQ, \r \in \sR$ and is necessary in order to avoid overloaded queues. 
In practice housing resources are strictly preferred. As a result, we propose to study the system under the so-called \emph{heavy traffic} regime, where the system is loaded very close to its capacity and we assume that the system utilization parameter $\rho := \mu_{\sR}/\lambda_{\sQ}$ approaches 1, i.e., $\rho \rightarrow 1$. In general, we assume that $\bm \lambda$ and $\bm \mu$ are such that $\lambda_{\sQ} = \rho \mu_{\sR}$. This assumption will additionally make the analytical study of the matching system more tractable. In particular, in~\cite{Afeche2021OnSystem}, the authors propose a quadratic program to approximate the exact steady-state flows of the stochastic FCFS matching system
under heavy traffic conditions. They enforce the steady-state flows in an optimization model to find the optimal matching topology using KKT optimality conditions. We adopt the same set of constraints in Section~\ref{sec:optimization} where we present the optimization formulation. 
We let $\vF := [\f_{\q\r}] \in \mathbb{R}_{+}^{|\sQ| \times |\sR|}$ denote the steady-state flow, where $\m_{\q\r} = 0 \Rightarrow \f_{\q\r} = 0$. Given a partitioning function $\P$, the policy associated with a matching topology $\vM$ is equal to $\pi_{\P, \vM}(\r | \bm x) = \f_{\q\r}/\sum_{\r\in\sR}{\f_{\q\r}} = \f_{\q\r}/\lambda_{\q}$, in which $\q = \P(\vx)$ and the second inequality follows from the flow balance constraints.
In Proposition~\ref{prop:policy_value} we show how the policy value can be written using the matching model parameters and treatment effect of different interventions. We define the conditional average treatment effect (CATE) of resource $\r$ and queue $\q$ as $\tau_{\q\r} := \mathbb{E}[Y(\r) - Y(1) | P(\vX) = \q] \; \forall \r \in \sR, \q \in \sQ$, in which $\r = 1$ is the baseline intervention. {In many applications, the baseline intervention corresponds to ``no-intervention'' (also referred to as the control group)}. In the housing allocation context, we set $\r=1$ to be the SO intervention.

\begin{proposition}
Given a partitioning function $\P$, an MCMS model $(\sQ, \sR, \bm \lambda, \bm \mu, \vM)$, and the steady-state FCFS flow $\vF$ under FCFS discipline, the value of the induced policy $\pi_{\P,\vM}$ is equal to: 
\begin{equation*}
    V(\pi_{\P, \vM}) = \frac{1}{\lambda_{\sQ}}\sum_{\q \in \sQ} \sum_{\r \in \sR} \f_{\q\r}\tau_{\q\r} + C,
\end{equation*}
where 
$C$ is a constant that depends on the expected outcome under the baseline intervention.
\label{prop:policy_value}
\end{proposition}


\subsection{Policy Optimization}


We now introduce the policy optimization problem under the assumption that the joint distribution of $\bm X, Y(\r), \r \in \sR$ as well as the partitioning function $\P$ is known. The problem formulation is as follows: 
\begin{equation}
    \sP(\P) := \max_{\vM \in \sM} V(\pi_{\P, \vM}),
    \label{prob:policy-optimization}
\end{equation}
Here, $\sM$ is the set of admissible matchings and imposes steady-state flow, fairness and minimum wait time constraints.
\paragraph{Fairness}
In this work, we focus on group-based notions of fairness which have been widely studied in recent years in various data-driven decision making settings~\cite{Rahmattalabi2019ExploringProblems,Azizi2018DesigningResources,Nguyen2021ScarceJustice,Bertsimas2013FairnessTransplantation}. Formally, we let $G$ be a random variable describing the group that an individual belongs to, taking values in $\sG$. For example, $\G$ can correspond to protected features such as race, gender or age. It is also possible to define fairness with respect to other features, such as vulnerability score in the housing allocation setting. We give several examples to which our framework applies.
\begin{example}[Maximin Fairness in Allocation]
Motivated by Rawls theory of social justice~\cite{Rawls1999TheoryEdition}, maximin fairness aims to help the worst-off group as much as possible. Formally, the fairness constraints can be written as
\begin{equation*}
    \sum_{\q \in \sQ_\g} \f_{\q\r} \geq w \quad \forall \g \in \sG, \r \in \sR,
\end{equation*}
where $w$ is the minimum acceptable flow across groups and $\sQ_\g \subseteq \mathcal Q$ is a subset of queues whose individuals belong to $\G = \g$. If queues contain individuals with different values of $g$, one should separate them by creating multiple queues with unique $g$. 
By increasing the parameter $w$, one is imposing more strict fairness requirements. This parameter can be used to control the trade-off between fairness and policy value. It can also be set to the highest value for which the constraint is feasible. 
\end{example}

\begin{example}[Group-based Parity in Allocation]
Parity-based fairness notions strive for equal outcomes across groups. 
\begin{equation*}
    \left|\sum_{\q \in \sQ_\g} \f_{\q\r} - \sum_{\q \in \sQ_{\g'}} \f_{\q\r}\right| \leq \epsilon \quad \forall \g, \g' \in \sG, \r \in \sR.
\end{equation*}
In words, for every resource the difference between the cumulative flow between any pair of groups should be at most $\epsilon$,
where $\epsilon$ can be used to control the trade-off between fairness and policy value.
\end{example}

\begin{example}[Maximin Fairness in Outcome]
For every group, the policy value should be at least $w$.
\begin{equation*}
 \frac{1}{\lambda_{\mathcal Q_g}}\sum_{\q \in \mathcal Q_g} \sum_{\r \in \sR} \f_{\q\r}\tau_{\q\r} \geq w \quad \forall \g \in \sG.
\end{equation*}
\end{example}

\begin{example}[Group-based Parity in Outcome]
The difference between the policy value for any pair of groups is at most $\epsilon$.
\begin{equation*}
     \left|\frac{1}{\lambda_{\sQ_\g}}\sum_{\q \in \sQ_\g} \sum_{\r \in \sR} \f_{\q\r}\tau_{\q\r} - \frac{1}{\lambda_{\sQ_{\g'}}}\sum_{\q \in \sQ_\g'} \sum_{\r \in \sR} \f_{\q\r}\tau_{\q\r} \right|  \leq \epsilon \quad \forall \g, \g' \in \sG.
\label{def:fairness-parity}
\end{equation*}
\end{example}
In the experiments, we focus on fairness in outcome due to treatment effect heterogeneity. In other words, it is important to match individuals with the right type of resource, rather than ensuring all groups have the same chance of receiving any particular resource. Further, we adopt maximin fairness which guarantees Pareto optimal policies ~\cite{Rahmattalabi2021FairApproach}.
\paragraph{Wait Time}
Average wait time is dependent on the structure of the matching topology. For example, minimum average wait time is attainable in a fully FCFS policy where $\m_{\q\r} = 1\; \forall \q \in \sQ, \r \in \sR.$ In~\cite{Afeche2021OnSystem}, the authors characterize the general structural properties that impact average wait time. In particular, they show that under the heavy traffic condition, a matching system can be partitioned into a collection of complete resource pooling (CRP) subsystems that operate ``almost'' independently of
each other. A key property of this partitioning is that individuals that belong to the same CRP component experience the same average steady-state wait time. Furthermore, the average wait time is tied to the number of CRPs of a matching topology, where a single CRP achieves minimum average wait time. In~\cite{Afeche2021OnSystem},  the authors introduce necessary and sufficient constraints to ensure that the matching topology $\vM$ induces a single CRP component. {We adopt these constraints in order to achieve minimum wait time which we discuss next.}
\subsection{Optimization Formulation}\label{sec:optimization}
Suppose the joint distribution of $X$, $Y(\r) \; \forall \r \in \sR$ is known. For a given $\P$,  problem~\eqref{prob:policy-optimization} can be solved via the MIO: 
\begin{subequations}
    \begin{align}
    \max & \displaystyle \sum_{\q \in \sQ}\sum_{\r\in\sR} \tau_{\q\r}\f_{\q\r} & &
    \\
    \text{s.t. } & \displaystyle \f_{\q\r},\nu_{\q\r} \in \mathbb R_{+}, \gamma_{r}, \theta_{q} \in \mathbb R \quad \forall \q \in \sQ, \r \in \sR &
    \\
    & \displaystyle \m_{\q\r}, z_{\q\r} \in \{0,1\} \quad  \forall \q \in \sQ, \r \in \sR &
    \\
    & \displaystyle g^{(k)}_{\q\r} \in \mathbb{R}_{+} \quad \forall \q, k \in \sQ, \r \in \sR & 
    \\
    & \displaystyle \sum_{\q\in\sQ} \f_{\q\r} = \mu_{r} \quad \displaystyle \forall \r \in \sR & \label{eq-flow-1}
    \\
    & \displaystyle \sum_{\r\in\sR} \f_{\q\r} = \lambda_{\q} \quad \displaystyle \forall \q\in\sQ & \label{eq-flow-2}
    \\
    & \displaystyle \f_{\q\r} \leq \lambda_{q}\mu_{\r}\left(\theta_{\q}+\gamma_{\r}+\nu_{\q\r}\right) + Z(1-m_{\q\r}) \quad \displaystyle \forall \q \in \sQ, \r \in \sR & \label{eq-kkt-1}
    \\
    & \displaystyle \f_{qr} \geq \lambda_{q}\mu_{r}\left(\theta_{q}+\gamma_{r}+\nu_{\q\r}\right)-Z(1 - m_{qr}) \quad \forall q\in\sQ, r\in\sR & \label{eq-kkt-2}
    \\
    & \displaystyle \f_{qr} \leq Z m_{qr} \quad \forall q \in \sQ, r\in\sR & \label{eq-kkt-3}
    \\
    & \displaystyle \f_{qr} \leq Z z_{qr} \quad \forall q\in\sQ, r\in\sR \label{eq-kkt-4}
    \\
    & \displaystyle \nu_{\q\r} \leq (|\sQ|+|\sR|+1)W(1-z_{qr}) \quad \forall q\in\sQ, r\in\sR \label{eq-kkt-5}
    \\
    & \displaystyle\sum_{\q \in \mathcal C} g_{qr}^{(k)} = \mu_r \quad \forall r \in \sR, k \in \sQ \label{eq-crp-1}
    \\
    &  \displaystyle\sum_{\r \in \sR} h_{qr}^{(k)} = \lambda_q - \frac{\delta}{|\sQ| - 1} \quad \forall q \in \sQ\setminus\{k\}, k \in \sQ  \label{eq-crp-2}
    \\
    &  \displaystyle\sum_{\r \in \sR} g_{qr}^{(k)} \leq Z m_{qr} \quad \forall q, k \in \sQ, r \in \sR \label{eq-crp-3}
    \\
    &  \displaystyle\sum_{r \in \sR}g_{qr}^{(k)} = \lambda_{k} + \delta \quad \forall q, k \in \sQ & \label{eq-crp-4}
    \\ 
    &  \frac{1}{\lambda_{\sQ_g}}\sum_{q \in \sQ_g} \sum_{r \in \sR} f_{qr}\tau_{qr} + C \geq w \quad \forall g \in \mathcal G
    , \vF \in \sF. & \label{eq-fairness}
\end{align}
In this formulation, $\delta := \left(\prod_{\q \in \sQ}v_{\q}\prod_{\r \in \sR}v_{\r}\right)^{-1}$ and $\frac{w_{\q}}{v_{\q}} = \lambda_{\q}, \frac{w_{\r}}{v_{\r}} = \mu_{\r}$ are rational number representations. Further, $W := 1/2 \max \{\max_{\q\in\sQ} 1/\lambda_{\q},\max_{\r\in\sR} 1/\mu_{r}\}$, and $\textstyle Z := \lambda^{\text{(max)}} \mu^{\text{(max)}} \left(\sum_{\q\in\sQ} 1/\lambda_{\q} + \sum_{\r\in\sR} 1/\mu_{\r} + (|\sQ|+|\sR|+1)^2 W \right)$, where $\lambda^{\text{(max)}}$ and $\mu^{\text{(max)}}$ are the maximum arrival rate across demand (arriving individuals) and resource queues, respectively.
\label{prob:MILP}
\end{subequations}
Constraints \eqref{eq-flow-1} and \eqref{eq-flow-2} are the flow balance constraints. Constants $W, Z$ ensure that constraints~\eqref{eq-kkt-1}-\eqref{eq-kkt-5} impose the KKT conditions of the quadratic program that approximates steady-state-flow for a matching topology $\vM$. Constraints~\eqref{eq-crp-1}-\eqref{eq-crp-4} enforce a single CRP component to ensure minimum wait time. 
Finally, constraint~\eqref{eq-fairness} collects the fairness constraints where we can use any of the aforementioned examples. In order to solve problem~\eqref{prob:MILP}, we need to estimate $\tau_{\q\r}$ and $\bm \lambda$ which depend of $\P$, as well as $\bm \mu$. 
\newar{Once the queues are fixed, estimating $\bm \lambda$ values are straightforward. In addition, $\bm \mu$ can be easily estimated from the historical provision of resources. In the next section, we discuss how to construct the queues and simultaneously estimate the $\tau_{\q\r}$ values.}

\section{Solution Approach}\label{sec:solution}

We first partition $\sX$ and then estimate CATE in each subset of the partition. We propose to use causal trees to achieve both tasks simultaneously~\cite{Wager2018EstimationForests}. Causal trees estimate CATE of binary interventions by partitioning the feature space into sub-populations that differ in the magnitude of their treatment effects. The method is based on regression trees, modified to estimate the goodness-of-fit of treatment effects. A key aspect of using causal trees for partitioning is that the cut points on features are such that the treatment effect variance within each leaf node is minimized. In other words, individuals who are similar in the treatment effect are grouped together in a leaf node. This results in queues that are tied to the treatment effect of resources which will result in improved policy value (see Section~\ref{sec:experiments}). 
\subsection{Assumptions}
Causal trees rely on several key assumptions which are standard in causal inference for treatment effect estimation~\cite{Hernan2013CausalBook}. These assumptions are {usually} discussed for the case of binary treatments. Below, we provide a generalized form of the assumptions for multiple treatments. 

\begin{assumption}[Stable Unit Treatment Value Assumption (SUTVA)]
The treatment that one unit (individual) receives does not change the potential outcomes of other units. 

\end{assumption}
\begin{assumption}[Consistency]
The observed outcome agrees with the potential outcome under the treatment received.
\end{assumption}
The implication of this assumption is that there are no different forms of each treatment which lead to different potential outcomes. In the housing allocation setting, this requires that there is only one version of PSH, RRH and SO.
\begin{assumption}[Positivity]
For all feature values, the probability of receiving any form of treatment is strictly positive{, i.e.,}
$$
\mathbb{P}(\R = \r |\vX = \vx) > 0 \; \forall \r \in \sR \; \vx \in \sX.$$
\end{assumption}
Th positivity assumption states that any individual should have a positive probability of receiving any treatment. Otherwise, there is no information about the distribution of outcome under some treatments and we will not be able to make inferences about it. In Section~\ref{sec:experiments}, we discuss the implications of this assumption in the context of HMIS data.
\begin{assumption}[Conditional Exchangeability]
Individuals receiving a treatment should be considered exchangeable, with respect to potential outcomes $Y(\r), \r \in \sR$, with those not receiving it and vice versa. Mathematically, 
$$
Y(1), \dots, Y(|\sR|) \perp R |\vX = \vx \; \forall \vx \in \sX.
$$
\end{assumption}
Conditional exchangeability means that there are no unmeasured confounders that are a common cause of both treatment and outcomes. If unmeasured confounders exist, it is impossible to accurately estimate the causal effects. 
In observational settings, a decision-maker only relies on passive observations. As a result, in order to increase the plausibility of this assumption, researchers typically include as many features as possible in $\vX$ to ensure that as many confounders as possible between treatment and outcome are accounted for. In the housing allocation setting, the HMIS data contains a rich set of features (54 features) associated with different risk factors for homelessness which we use in order to estimate the treatment effects. In Section~\ref{sec:discussion}, we discuss the consequences of violating the above assumptions.

\subsection{Building the Partitioning Function}
Next, we describe our approach for estimating CATE. 
We first consider a simple case with binary treatments, i.e., $|\sR| = 2$ as causal trees work primarily for binary treatments.
After training the causal tree using the data on a pair of treatments, the leaves induce a partition on the feature space $\sX$. Hence, we can view the causal tree as the partitioning function $\P$ where each individual is uniquely mapped to a leaf node, i.e., a queue.

Extending to the case of $|\sR| > 2$ is non-trivial. 
Assuming $\r = 1$ is the baseline intervention, we  construct $|\sR|-1$ separate causal trees to estimate CATE for $\r \in \sR\setminus\{1\}$. We denote the resulting causal trees or partitioning functions $\P_{\r}:\sX \rightarrow \sQ \; \forall \r \in \sR\setminus\{1\}$. We define $\sX_{\r}(\q) = \left\{ \vx \in \sX: \P_{\r}(\vx) = \q \right\} \forall \r \in \sR\setminus\{1\}, \q \in \sQ$ as the set of all individuals who belong to queue $\q$ according to partitioning function $\P_r$. Also, let $\q_{\r} = \P_{\r}(\vx)$.
\begin{figure}
  \begin{center}
        \includegraphics[width=0.4\textwidth]{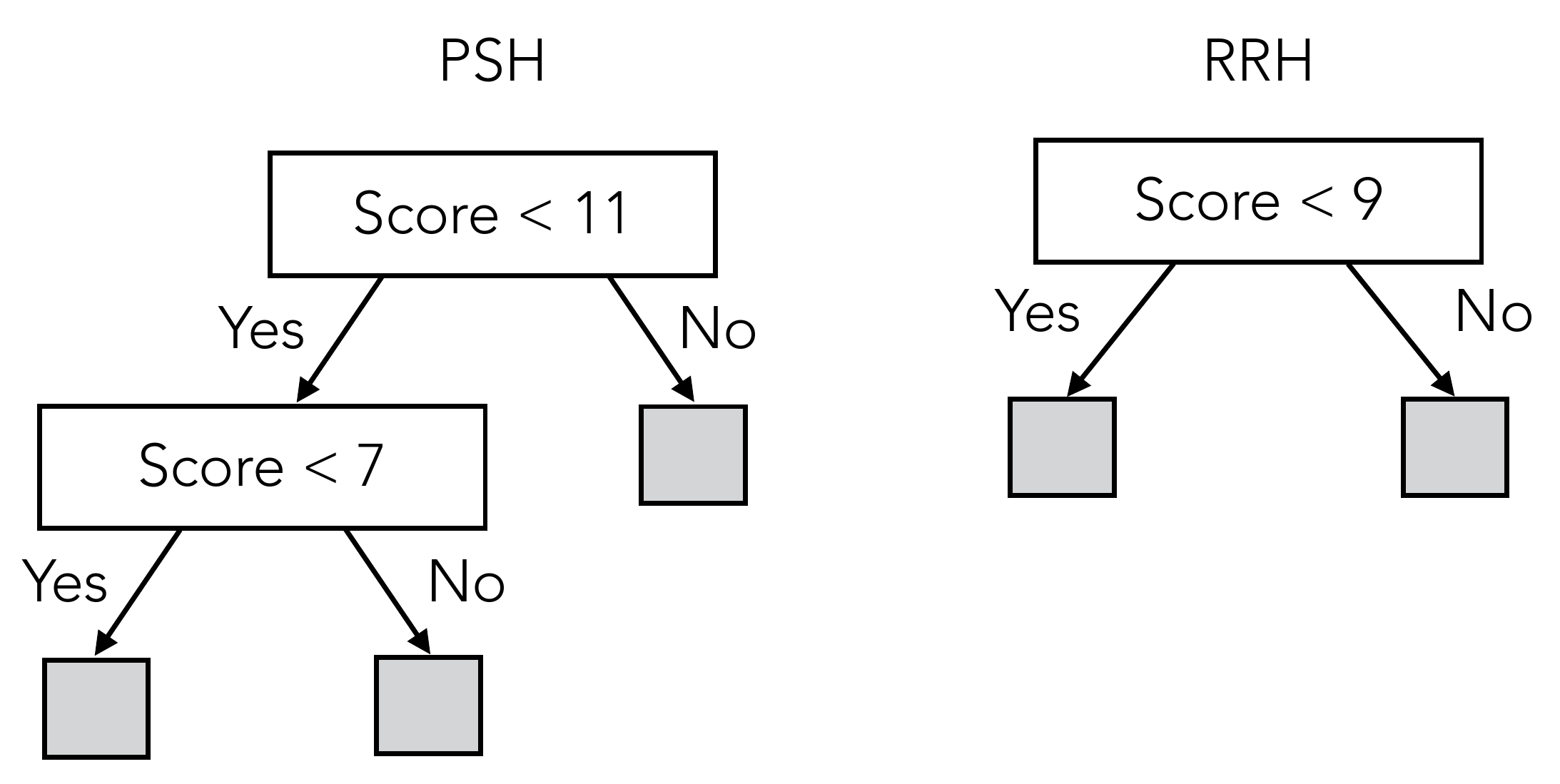}
  \end{center}
  \caption{Example partitioning by sample causal trees for PSH and RRH interventions.}
  \label{fig:example}
\end{figure}
In order to aggregate the individual partitioning functions to obtain a unified partition on $\sX$, we consider the intersections of $\sX_{\r}(\q)$ created by each tree. We define subsets $\sX(\q_{1}, \dots, \q_{|\sR|-1}) = \bigcap^{|\sR|-1}_{\r = 1} \sX_{\r}(\q_{\r})$ for all combinations of $\q_\r \in \sQ$. We can view $\sX(\q_{1}, \dots, \q_{|\sR|-1})$ as a new (finer) partition on $\sX$.
We illustrate with an example using the housing allocation setting. Suppose we have constructed two causal trees for PSH and RRH according to Figure~\ref{fig:example} such that PSH tree splits the vulnerability score into intervals of $[0,6], [7, 10], [11, 17]$ and RRH creates $[0, 8], [9, 17]$ subsets. According to our procedure, the final queues are constructed using the intersection of these subsets. In other words, we obtain $[0, 6], [7, 8], [9, 10], [11, 17]$ which corresponds to four queues. \newar{Building queues as the intersection of the tree subsets may result in a large number of queues. However, even though the size of the optimization problem is quadratic in the number of queues, in practice the problem solves relatively quickly (in seconds) for queue number up to 30 which is far more than what is typically seen in practice. In addition, the granularity of the partition, and subsequently the number of queues, can be controlled through the tree depth or the minimum allowable number of data points in each leaf, both of which are adjustable parameters in causal trees. It is noteworthy that while fewer queues result in models that are easily interpretable, more queues allow the decision-makers to leverage the heterogeneity in treatment effect to target the resources to the right individuals, hence achieving higher-valued policies. We explore this trade-off in the experimental result section.}


Finally, in order to estimate $\tau_{\q\r}$, we should avoid using the estimates from each individual tree.  
The reason is that each tree estimates $\mathbb{E}[Y(\r) - Y(1) |\P(\vX) = \q, \R \in \{1, \r\}] \; \forall \r \in \sR\setminus\{1\}$. That is, a subset of the data associated with a pair of treatments is used to build each tree. Therefore, $\tau_{\q\r}$ values are not generalizable to the entire population and need to be re-evaluated over all data points that belong to a subset. 
We adopt Doubly Robust estimator (DR) for this task. Proposed in~\cite{Dudik2011DoublyLearning}, DR combines an outcome regression with a model for the treatment assignment (propensity score) to estimate treatment effects. DR is an unbiased estimate of treatment effects, if at least one of the two models are correctly specified. Hence, it has a higher chance of reliable inference. CATE estimates $\hat{\tau}_{\q\r}$ are provided below.
\begin{equation*}
    \hat{\tau}_{\q\r} = T(\r) - T(1) \;  \r \in \sR, 
\end{equation*}
where $T(r) = \frac{1}{|\mathcal I_{\q}|}\sum_{i\in \mathcal I_{\q}}\left(\hat{y}(\bm X_i, r) + \left(Y_i - \hat{y}(\bm X_i, R_i)\right)\frac{\mathbb{I}(\R_i = r)}{\bar{\pi}(\R_{i}|\vX_i)}\right)$ and $\mathcal I_{\q} := \{i \in \{1,\dots,N\}: \P(\vX_{i}) = \q\}$ is the set of indices in the historical data that belongs to $\q$. Further, $\hat{y}$ and $\bar{\pi}$ are the outcome and historical policy (i.e., propensity score) models, respectively. According to the above expression, all resources are compared to the baseline intervention, hence $\hat{\tau}_{\q\r} = 0$ for $\r = 1$.

We end this section by discussing a practical consideration which is a desire to design policies that depend on low-dimensional features, such as risk scores. In cases that we only use risk scores, not the full feature vector, it is critical that they satisfy the causal assumptions. We provide a risk score formulation that satisfies this requirement.
\begin{proposition}
We define risk score functions as $S_r(\vx) = \mathbb{P}[Y(\r) = 1 |\vX = \vx] \; \forall \r \in \sR$. Suppose $\bm S \in \mathcal S$ is a (random) vector
of risk scores. Also, let $\bm Y = (Y(1),\dots,Y(|\sR|))$ be the vector of potential outcomes. The following statements hold for all $\forall \vx \in \sX, \bm s \in \mathcal S$:
\begin{enumerate}
    \item $\bm Y \perp \R |\vX \Rightarrow \bm Y \perp \R |\bm S$. 
    \item $\mathbb{P}\left( \mathbb{P}(\R = \r |\vX = \vx) > 0 \right) = 1 \Rightarrow \mathbb{P}\left( \mathbb{P}(\R = \r |\bm S = \bm s) > 0 \right) = 1.$
\end{enumerate}
\label{prop:feature-summarization}
\end{proposition}
Under causal assumptions, $S_r(\vx) = \mathbb{P}(Y(\r) = 1 |\vX = \vx, \R = \r) = \mathbb{P}(Y = 1 |\vX = \vx, \R = \r)$, which relies on observed data, rather than counterfactuals. According to Proposition~\ref{prop:feature-summarization}, as in general individuals respond differently to various treatments, one risk score per resource may be required in order to summarize the information of $\vX$. \newar{Alternatively, one can utilize the entire set of features in the causal tree and to learn the propensity and outcome models used in the treatment effect estimators.}

\section{Computational Results}\label{sec:experiments}
We conduct two sets of experiments to study the performance of our approach to design resource allocation policies: \emph{(i)} synthetic experiments where the treatment and potential outcomes are generated according to a known model; \emph{(ii)} experiments on the housing allocation system based on HMIS data for youth experiencing homelessness. We use the causal tree implementation in the \texttt{grf} package in R. We control the partition granularity by changing the \emph{minimum node size} parameter which is minimum number of observations in each tree leaf. We evaluate policies using three estimators from the causal inference literature~\cite{Dudik2011DoublyLearning}:  Inverse Propensity Weighting (IPW) which corrects the mismatch 
between the historical policy and new policy by re-weighting the data points with their propensity values, Direct Method (DM) which uses regression models to estimate the unobserved outcomes, and DR. In addition, we include objective value of Problem~\eqref{prob:MILP} obtained by matching flow and CATE estimates (CT). When models of outcome and propensity are correctly specified, the above estimators are all unbiased~\cite{Dudik2011DoublyLearning}.  
\begin{figure}[t]
    \begin{subfigure}[b]{0.35\textwidth}
    \centering
         \includegraphics[width=\textwidth]{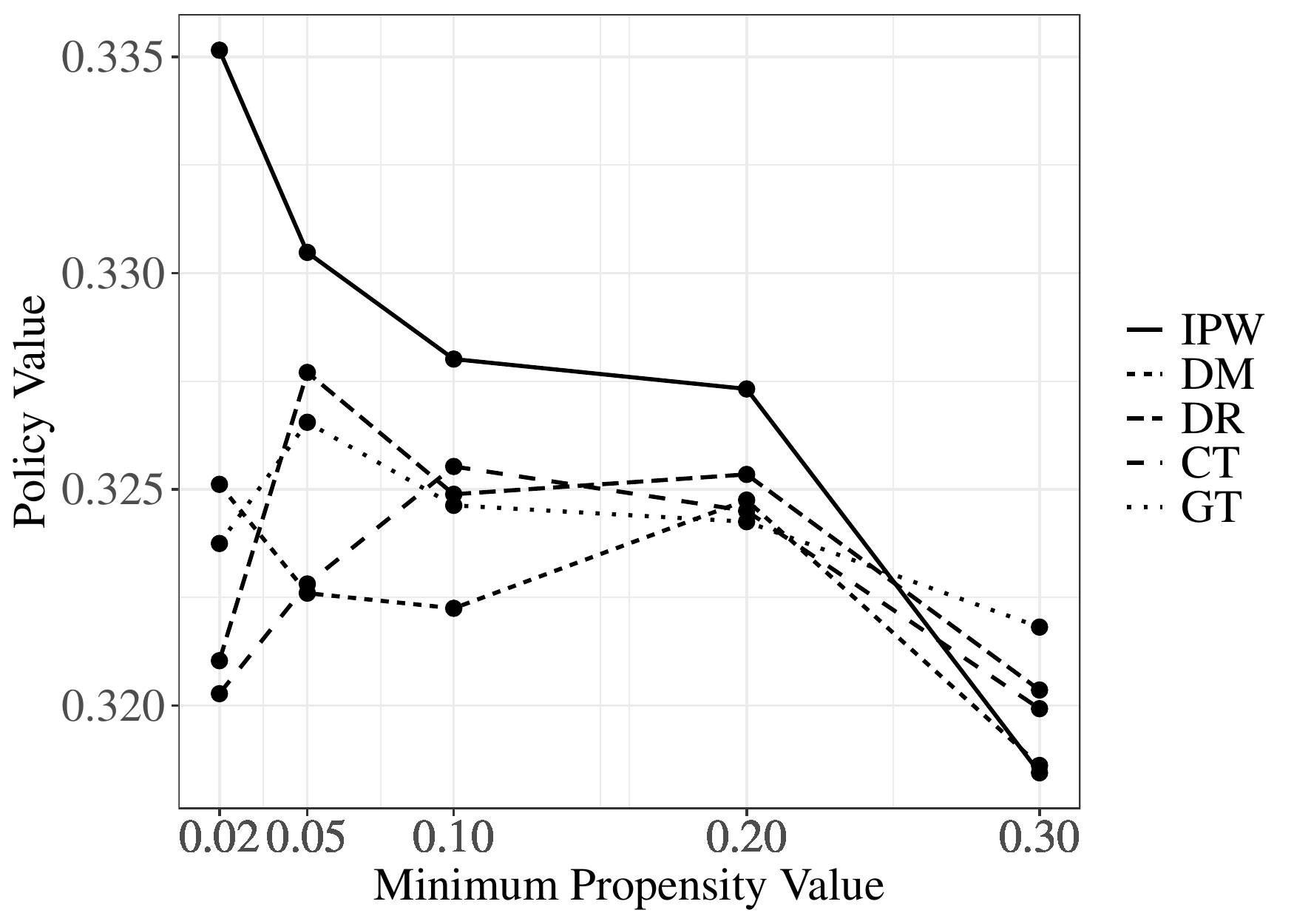}
         \caption{Policy value vs. the minimum propensity value.}
         \label{fig:small-prop}
    \end{subfigure}\qquad
    \begin{subfigure}[b]{0.35\textwidth}
    \centering
         \includegraphics[width=\textwidth]{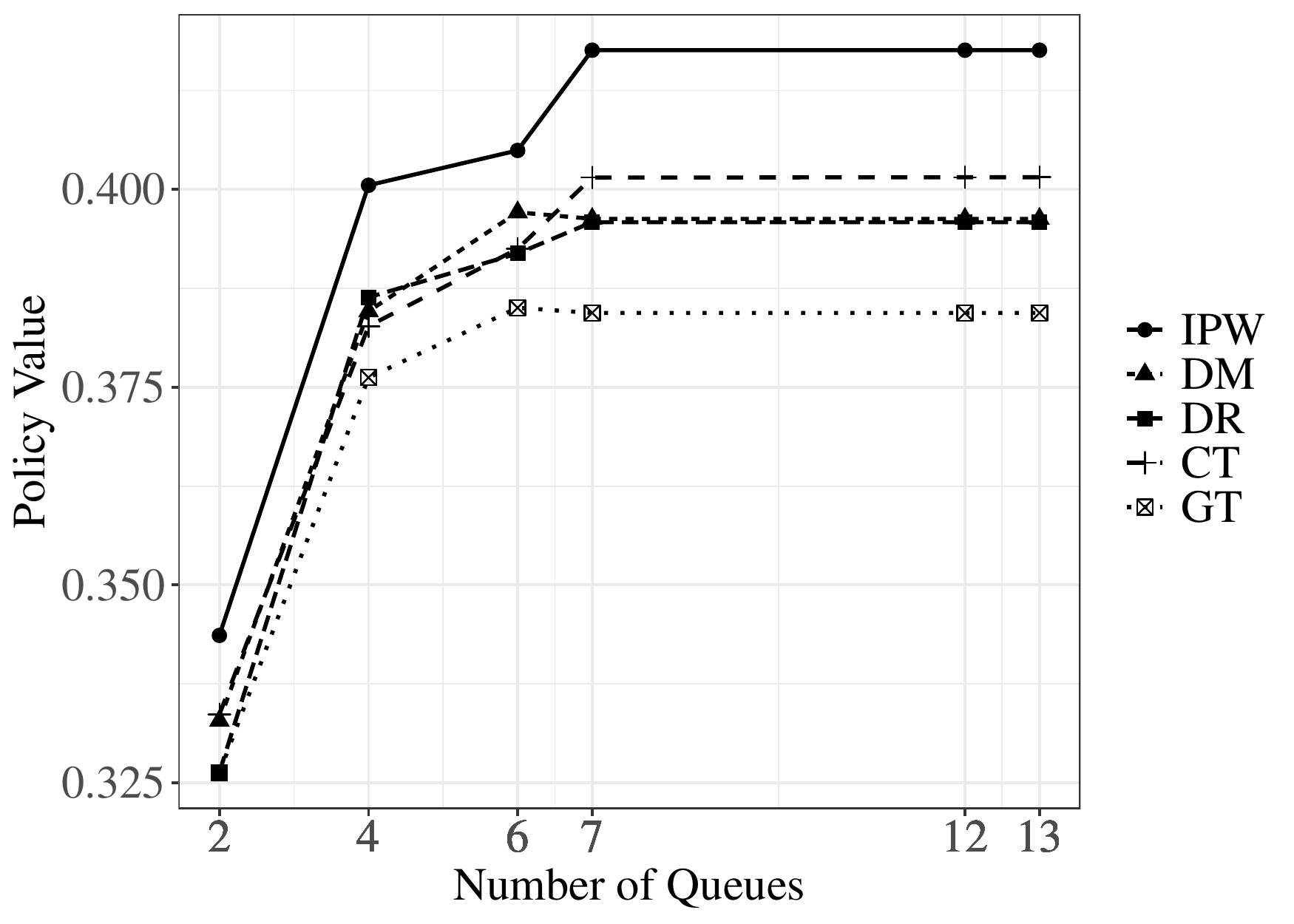}
         \caption{Policy value vs. the number of queues.}
         \label{fig:heter}
    \end{subfigure}  
    \caption{Synthetic data.
Each line corresponds to a different estimator. }
\vspace*{-\baselineskip}
\end{figure}
\subsection{Synthetic Experiments}
We generate synthetic potential outcomes and resource assignments in the HMIS data collected between 2015 and 2017 from 16 communities across the United States~\cite{Chan2017EvidenceYouth}. We use the following setting using vulnerability score $S$ (unless mentioned otherwise):
$\bar{\pi}(\text{SO} |S > 0.2) = 0.3$, $\bar{\pi}(\text{SO} |0.0 < S \leq 0.2) = 0.3$ and $\bar{\pi}(\text{SO} |S \leq 0.0) = 0.3$. Additionally, $\bar{\pi}(\text{RRH} |S > 0.2) = 0.2$, $\bar{\pi}(\text{RRH} |0.0 < S \leq 0.2) = 0.4$ and $\bar{\pi}(\text{RRH} |S \leq 0.0) = 0.3$ and finally, $\bar{\pi}(\text{PSH} |S > 0.2) = 0.5$, $\bar{\pi}(\text{PSH} |0.0 < S \leq 0.2) = 0.3$ and $\bar{\pi}(\text{PSH} |S \leq 0.0) = 0.4$. The potential outcomes are sampled from binomial distributions with probabilities that depend on $S$. For PSH, we use $\mathbb E[Y(\text{PSH}) |S \leq 0.3] = 0.6$, $\mathbb E[Y(\text{PSH}) |0.3 < S \leq 0.5] = 0.2$ and $\mathbb E[Y(\text{PSH}) |0.5 < S] = 0.6$. For RRH, $\mathbb E[Y(\text{RRH}) |S \leq 0.2] = 0.2$, $\mathbb E[Y(\text{RRH}) |0.2 < S \leq 0.7] = 0.6$ and $\mathbb E[Y(\text{RRH}) |0.7 < S] = 0.2$. Finally, $\mathbb E[Y(\text{SO})]= 0$. We evaluate policies obtained by solving  Problem~\eqref{prob:MILP}. We use decision trees for outcome and propensity score models.


One of the goals of the synthetic experiments is to compare different estimators in a setting where we observe the potential outcomes. Specifically, we study the performance of the estimators for policy evaluation when propensity values are varied. 
We generate different datasets by changing the  propensity values $\bar{\pi}(\text{PSH}|0.0 < S \leq 0.2) = \alpha$ and $\bar{\pi}(\text{RRH}|0.0 < S \leq 0.2) = 0.7 - \alpha$ for $\alpha \in \{0.02, 0.05, 0.1, 0.2, 0.3\}$ and obtain the optimal policy for each dataset. Figure~\ref{fig:small-prop} shows optimal policy values according to different estimators. We observe that across the $x-$axis range, DR, DM and CT result in similar estimates which also agrees with the ground truth (GT). However, when the minimum propensity score is small ($< 0.05$), IPW diverges from GT. This is consistent with other findings in the literature suggesting that when propensities are too close to 0 or 1, non-parametric estimators tend to have higher variance and converge at a slower rate (with the number of data points)~\cite{Khan2010IrregularEstimation}.

Next, we investigate the effect of treatment heterogeneity on the value of the optimal policy. In particular, we study how much the granularity of partitions, or the number of queues, impacts the policy value. Figure~\ref{fig:heter} summarizes the results. 
When the number of queues is equal to 1, the optimal policy is at its minimum value. In this case, the policy corresponds to an FCFS policy as individuals queue in a single line and are prioritized according to the their arrival times. The optimal policy value gradually increases ($\sim25\%$ according to GT) as the number of queues increases until it flattens. This suggests that by increasing the number of queues, we can leverage the treatment effect heterogeneity across the queues to allocate resources more efficiently.



\subsection{HMIS Data of Youth Experiencing Homelessness}
We now showcase the performance of our approach to design policies that allocate resource among the U.S. homeless youth. We defer the details on data preparation to the Appendix.

\subsection{Data Pre-Processing and Estimation}
\noindent\textit{Outcome Definition. }We focus on the likelihood of stable exit from homelessness. 
\begin{figure}[t]
    \begin{subfigure}[b]{0.35\textwidth}
    \centering
         \includegraphics[width=\textwidth]{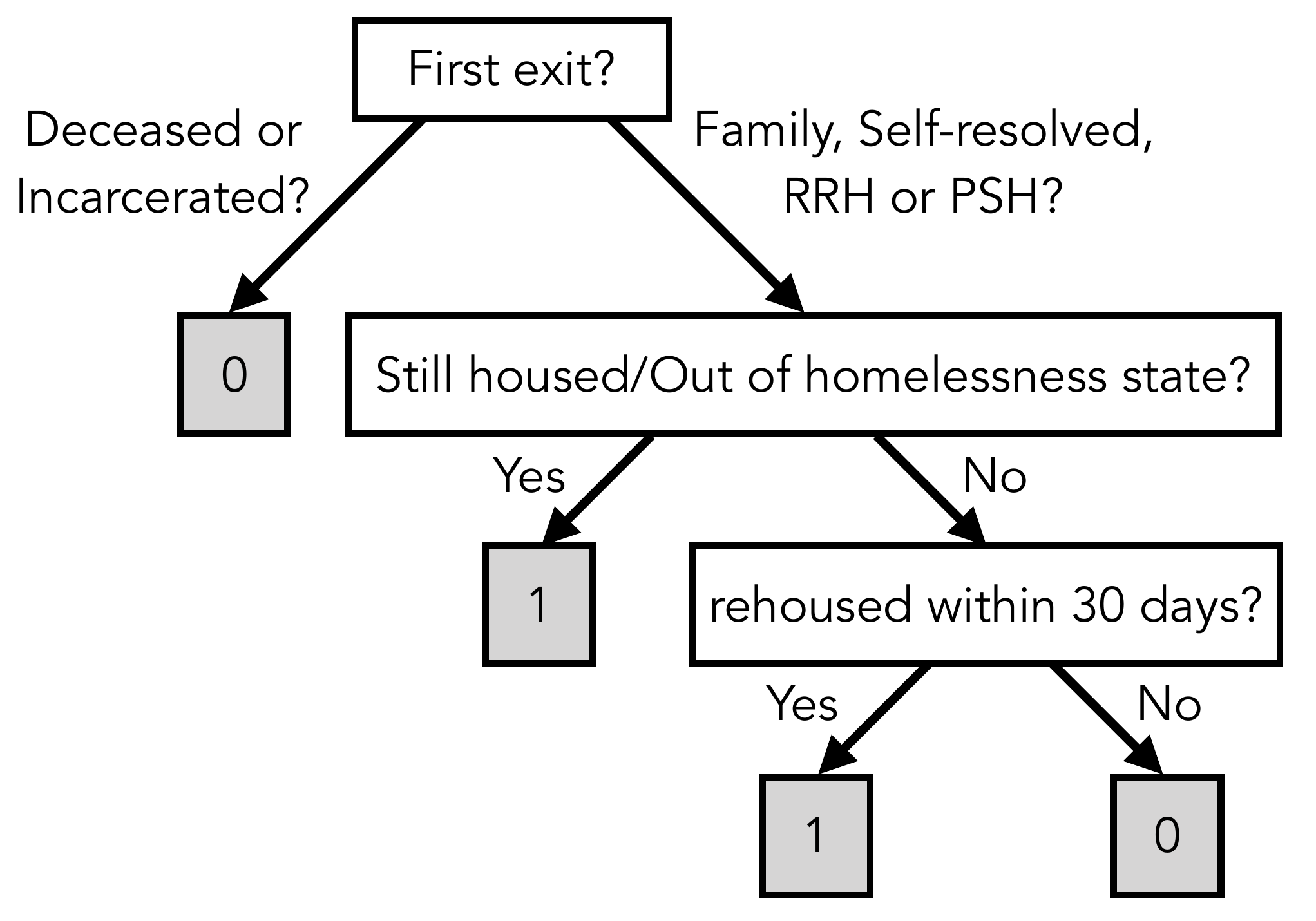}
         \caption{Success definition flow chart.}
         \label{fig:success-definition}
    \end{subfigure}\hfill
    \begin{subfigure}[b]{0.45\textwidth}
    \centering
         \includegraphics[width=\textwidth]{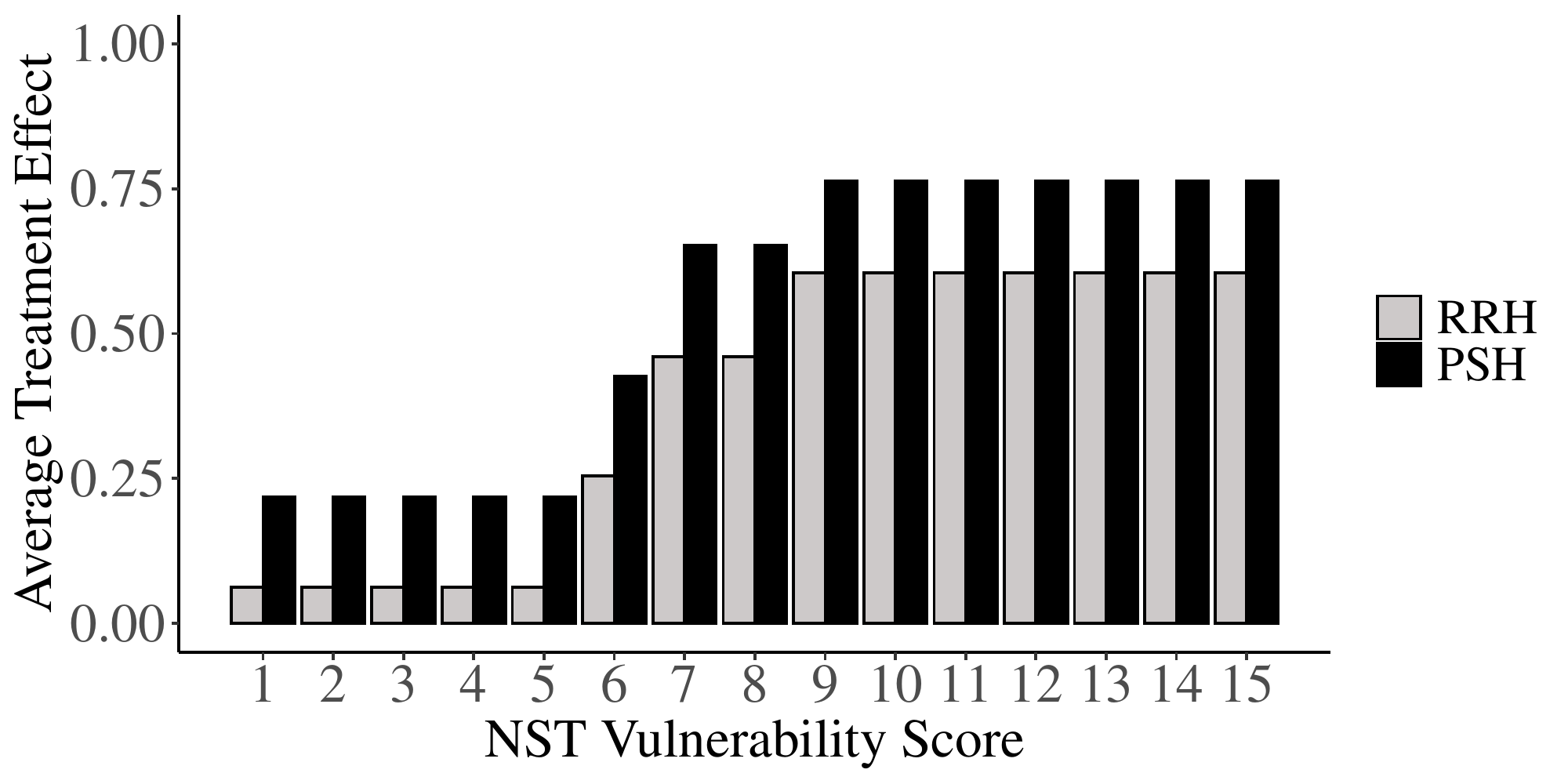}
         \caption{Heterogeneous treatment effect using DR method.}
         \label{fig:ceffect_score}
    \end{subfigure}   
    \caption{HMIS data.}
    \vspace*{-\baselineskip}
\end{figure}
An exit from the system can be to any of the following destinations: ``family,'' ``self-resolved,'' ``RRH,'' ``PSH,'' ``deceased,'' or ``incarcerated.'' Exiting due to incarceration or being deceased are undesirable outcomes and are encoded as $Y = 0$ (left branch). ``Family,'' ``self-resolve,'' ``RRH,'' and ``PSH'' are desirable outcomes but may be temporary exits, meaning that the individual may return to homelessness shortly after. In addition, there are recorded exits that are simply due to a ``move'' in the system from one service to another. 
We distinguish between these cases by checking whether an individual is ``still housed'', i.e., is at the stable exit destination. If re-housed, we consider a 30-day threshold to decide whether it is a return to homelessness ($Y = 0$) or a move in the system ($Y = 1$). This procedure for defining outcome is summarized in Figure~\ref{fig:success-definition}.

\noindent\textit{Propensity Estimation. }
In order to obtain an unbiased estimation of the policy value, IPW and DR approaches rely on propensity values. In our setting, the propensities are unknown but can be estimated from data. This poses a challenge to find a model that fits the data while being well-calibrated. 
We use different statistical models for multi-class classification
to estimate $\mathbb{P}(R = r |\bm X = \bm x), \forall \r \in \sR$. We evaluate models based on the predictive power, calibration, and fairness. For fairness, we adopt the test fairness criteria in~\cite{Chouldechova2017FairInstruments} since evaluating the policy value across different protected groups requires propensity values that are well-calibrated for those groups. We defer the details on model selection to the Appendix. 
We note that the original dataset does not satisfy the positivity assumption. That is, some groups of individuals have only received a subset of the resources. Therefore, for data points with propensities less than 0.001, we follow the status-quo policy and we exclude them from the policy optimization. 

\noindent\textit{Outcome Estimation. }
DM and DR methods rely on a model of the outcome under different resources. We compare an array of models in terms of accuracy, calibration, and test-fairness. The results are summarized in the Appendix.


\noindent\textit{Heterogenous Treatment Effect Estimation. }
We use causal trees with {minimum node size} {equal to} 15 to estimate the average treatment effects across the NST score range for RRH and PSH. According to Figure~\ref{fig:ceffect_score}, PSH consistently has a higher treatment effect than RRH indicating that it is a more effective resource. Further, the treatment effect of both resources increase with score which suggests that higher-scored individuals benefit more from these resources. We also provide results on the (unbiased) probability of exiting homelessness versus NST score in the Appendix. 

\noindent\textit{Arrival Rate. }
Once the queues are constructed, we estimate the arrival rate of individuals from data. Given the heavy-traffic condition, we calculate the required rate of SO as $\mu_{\text{SO}} = \max\left(\lambda_\sQ - \mu_{\text{RRH}} - \mu_{\text{PSH}}, 0\right)$. \newar{Further, in the HMIS data the resource arrival rates vary with time. In particular, between 2016 and 2017 there is a sharp decrease in the rate of PSH and RRH. Since the rate of resources is often known a-priori to the organisations, in the test data we re-evaluate the arrival rates and re-optimize according to those parameters.}

\subsection{Policy Optimization Results}

We now present the policy optimization results along three distinct objectives: policy value measured in terms of rate of stable exit from homelessness, fairness by race and age, and wait time. Table~\ref{tab:res-total} summarizes the results, where OPT is the optimal policy value without fairness constraints and OPT-fair (race), and OPT-fair (age) represent our method with fairness constraints over race and age, respectively. As baselines, we simulate both a fully FCFS policy and the status quo policy SQ (see Figure~\ref{fig:HMIS-matching}). We also compare with the deployed policy in the data SQ (data). As IPW suffers in small-propensity settings, we exclude it from the estimators.

\begin{table*}[t!]
    \centering
\begin{tabular}{P{2.0cm}P{1.5cm}P{1.5cm}P{1.5cm}P{3cm}}
\toprule
\multirow{2}{*}{Policy} & \multicolumn{3}{c}{Rates of Stable Exit from Homelessness} & \multirow{2}{*}{Wait Time (days)} \\
\cmidrule{2-4} & CT & DM & DR & \\
\hline
OPT & 0.76 & 0.74 & 0.75 & 142.67 \\
OPT-Fair (race) & 0.76 & 0.75 & 0.76 & 142.64 \\
OPT-Fair (age) & {0.76} & {0.75} & {0.75} & 142.64 \\
FCFS & 0.68 & 0.68 & 0.66 & 142.64 \\
SQ & 0.66 & 0.63 & 0.63 & 182.21 \\
SQ (data) & 0.73 & 0.73 & 0.73 & 156.77 \\
\bottomrule
\end{tabular}
\caption{{Out-of-sample estimated policy value} {measured} in terms of rates of stable exit from homelessness and wait times.}
\label{tab:res-total}
\vspace*{-\baselineskip}
\end{table*}
From Table~\ref{tab:res-total}, OPT, OPT-fair {(race), and OPT-far (age)} all outperform the baseline policies. Specifically, OPT significantly improves the rate of stable exit from homelessness by 19\% and 13\% (under DR estimates) over SQ and FCFS policies, respectively. Perhaps surprisingly, SQ performs worse than FCFS which is due to how the cut scores are designed. According to SQ individuals with scores 4-7 are matched to RRH. However, the RRH treatment effect is highest for scores above 7 (See Figure~\ref{fig:ceffect_score}). Compared to SQ (data), our policy values are competitive. We improve the wait time over SQ and SQ (data) by 21\% and 9\%, respectively and obtain values similar to FCFS policy. \newar{This is because we have imposed constraints that ensure a single CRP component and subsequently minimum wait time}. As a result, further algorithmic improvement is not possible unless problem inputs, such as resource arrival rates, change. \newar{We note that it is possible to relax the wait time constraints such that the average wait time is not minimum~\cite{Afeche2021OnSystem}. In our setting, this did not lead to an improvement in the policy value.} 
\begin{figure}[t!]
    \begin{subfigure}[t!]{0.4\textwidth}
    \centering
         \includegraphics[width=\textwidth]{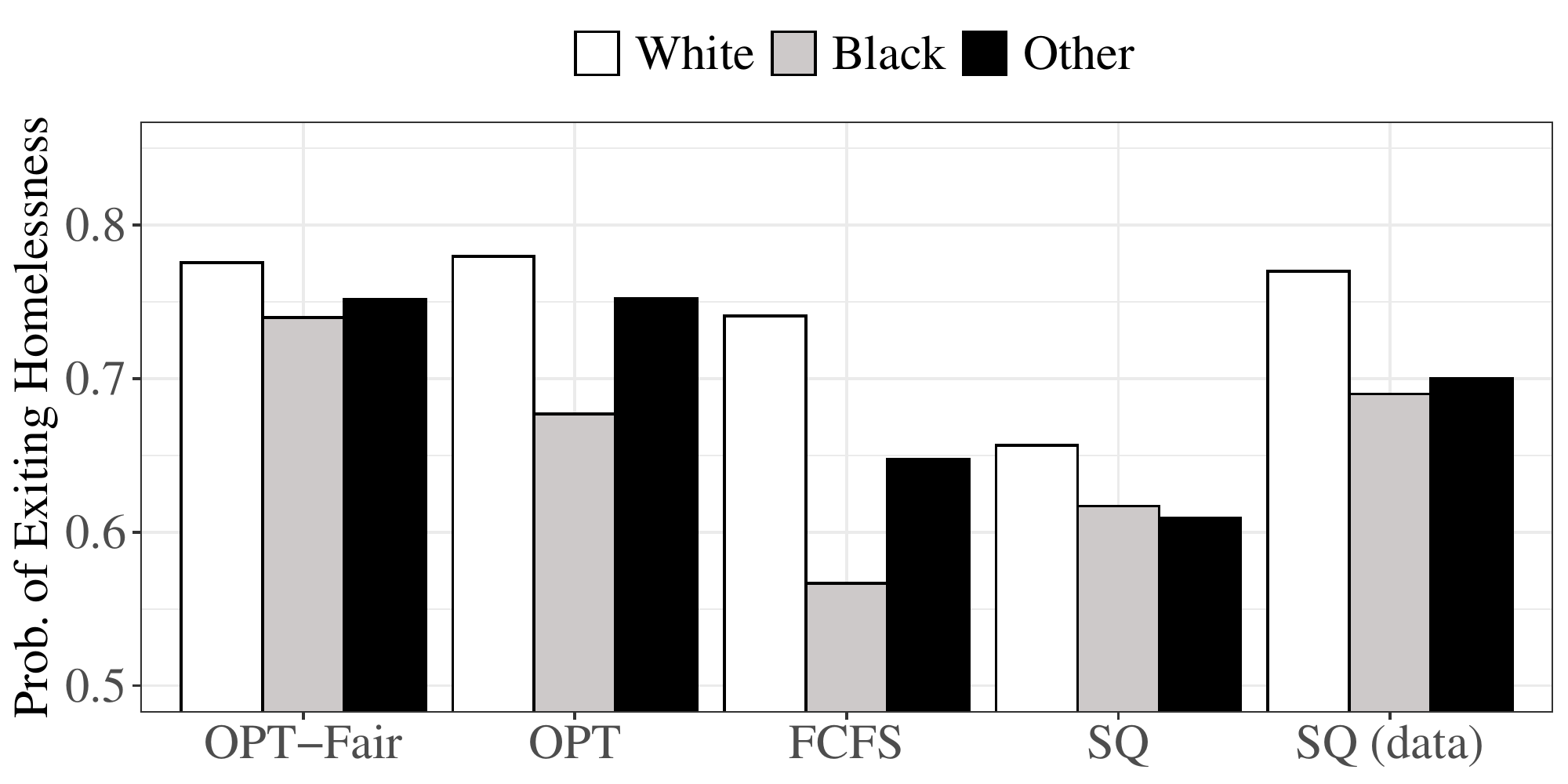}
    \end{subfigure}\qquad
    \begin{subfigure}[t!]{0.4\textwidth}
    \centering
         \includegraphics[width=\textwidth]{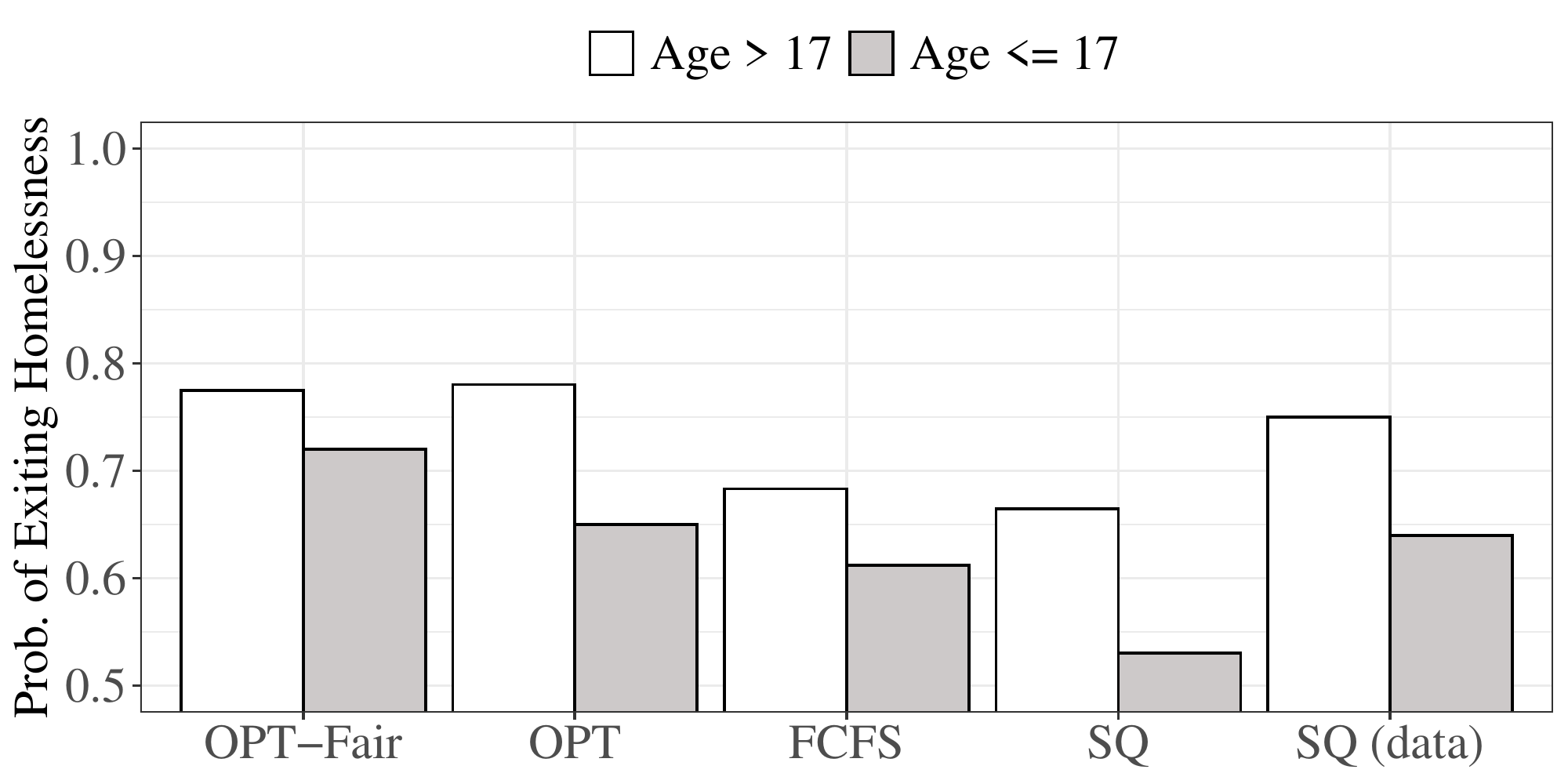}
    \end{subfigure}  
    \caption{Out-of-sample rates of exit from homelessness {by} race (left panel) and age (right panel) using the DR estimator.}
    \label{fig:outcome-fair}
    \vspace*{-\baselineskip}
\end{figure}
\begin{figure}[b]
\centering
    \includegraphics[width=0.25\textwidth]{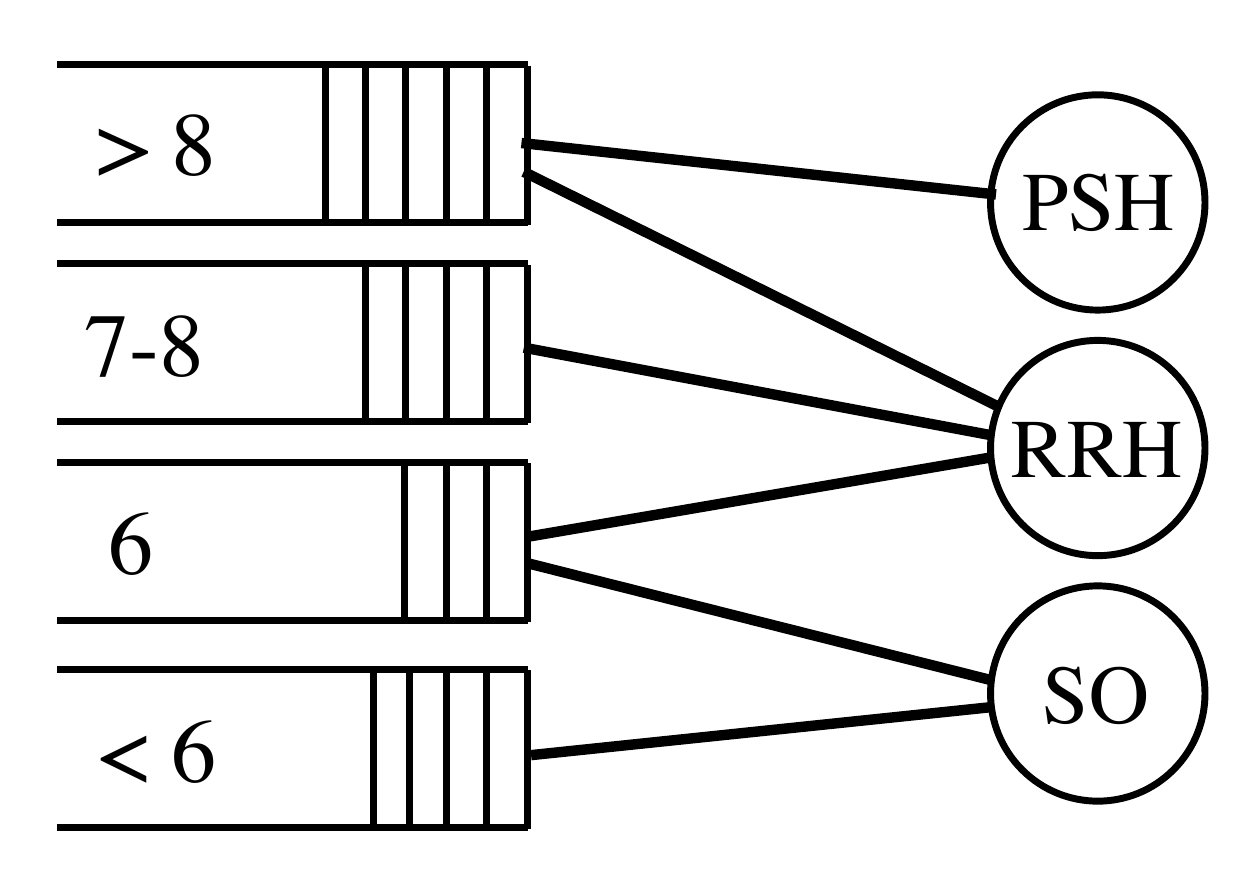}
    \caption{Optimal Topology}
    \label{fig:opt-topo}
\end{figure}
\newar{Finally, we observe that OPT and OPT-fair have similar policy values, indicating a small cost for fairness. This can be explained by the fact that OPT-fair optimizes over queues split by both the protected attribute and risk score, which provides more flexibility to target the resources to different protected groups despite the problem being more constrained.}
Figure~\ref{fig:outcome-fair} compares the worst-case rate of exiting homelessness across age (below and over 17 years old) and racial groups (White, Black, and Other) according to DR estimator and in the test data. First, we observe that an FCFS policy does not necessarily result in 
policies that are fair in terms of their outcomes neither by age nor by race. 
This is because FCFS policies ignore treatment effect heterogeneity. 
In other words, according to {the} FCFS discipline, everyone has the same probability of receiving any {one} of the resource types (fairness in allocation). However, not everyone benefits equally from the resources. Indeed, Black individuals seem to suffer the most under a fully FCFS policy. SQ also yields a low worst-case performance mainly due its low overall performance. SQ (data) has relatively {better} worst-case performance. However, there is still a significant gap between the performance of Black/Other groups and Whites. By explicitly imposing fairness constraints on policy outcomes across protected groups, OPT-Fair significantly improves the performance for the Black and Other groups. Figure~\ref{fig:outcome-fair}, similar observations can be made for fairness by age, where compared to baselines, OPT-Fair exhibits significant improvements in the policy value for those with age below 17.

\begin{figure}   \includegraphics[width =0.35\textwidth]{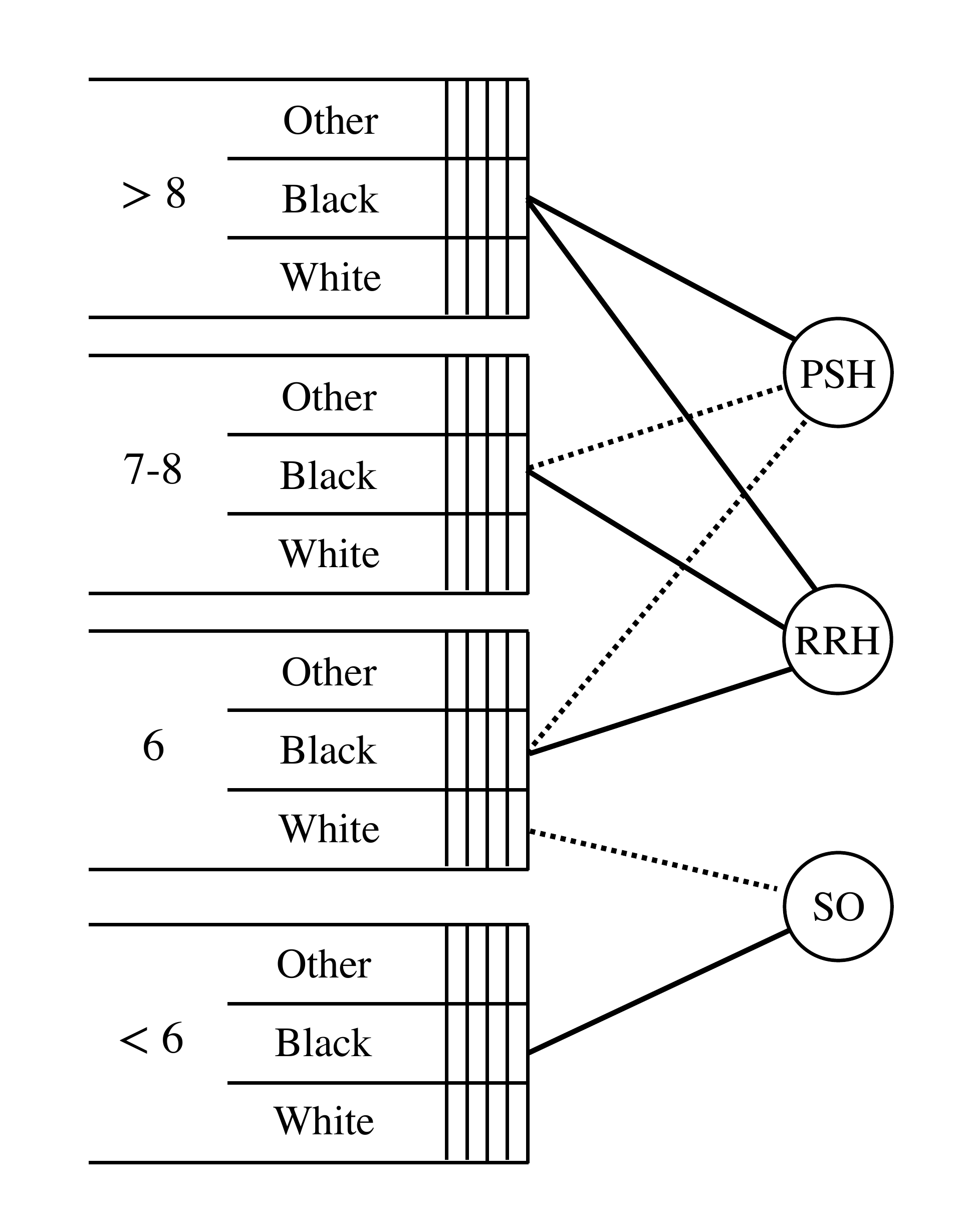}
    \caption{Optimal and fair matching topology by race.
    Individuals are divided into four different score groups: $S < 6$, $S = \{6\}$, $S = \{7,8\}$, $S > 8$. Queues are constructed based on score groups and race jointly.
    Solid lines indicate that a resource is connected to the entire score group (a collection of queues). Dotted lines indicate connection to a single queue within the score group. For example, SO is only connected to the individuals with $S = \{6\}$ and race White.}
    \label{fig:opt-race}
\end{figure}

We now present a schematic diagram of OPT and OPT-fair matching topologies. Figure~\ref{fig:opt-topo} is the matching topology corresponding to OPT policy. Compared to SQ, OPT uses different cut points on NST score, specifically for the lower-scoring individuals. Across the four score groups, we observe a gradual transition from eligibility for a more resource-intensive intervention (PSH) to a basic intervention (SO). Figure~\ref{fig:opt-race} depicts OPT-fair topology for fairness on race, in which queues are constructed using the joint values of NST score and race. According to this figure, PSH is matched to all individuals with scores above 9 as well as mid-scoring Black individuals, i.e., $6 \leq \text{score} \leq 9$. RRH is connected to every individual in the mid-score range. Our modeling strategy uses the protected characteristics in order to ensure fairness. This is motivated by discussions with our community advisory board, including housing providers/matchers and people with past history of homelessness, who suggested that in order to create a fair housing allocation system there ought to be special accommodations for historically disadvantaged people. 
\begin{figure}
    \centering
    \includegraphics[width=0.25\textwidth]{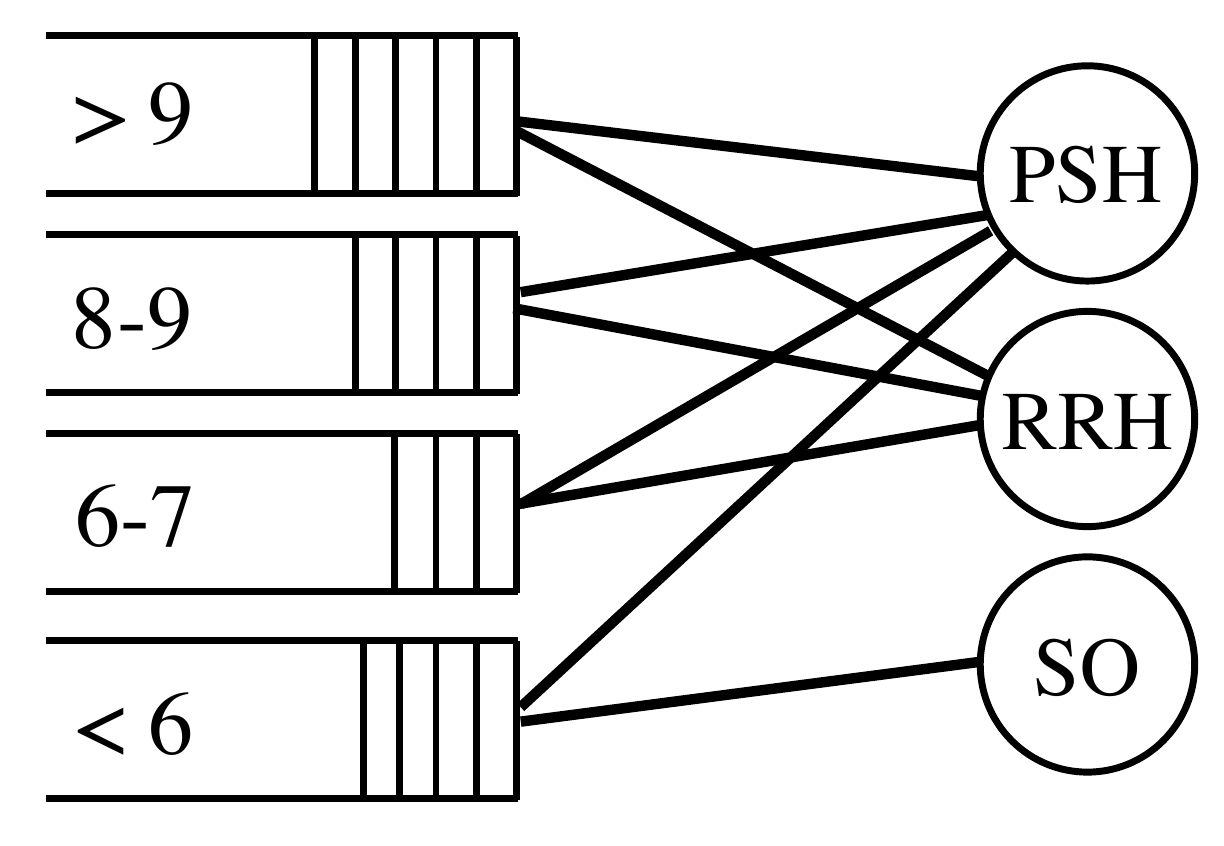}
    \caption{Fair topology when queues are not divided by race.}
    \label{fig:race-topo}
\end{figure}
Our policies align with affirmative action policies that take individuals' protected attributes into account in order to overcome present disparities of past practices, policies, or barriers by prioritizing resources for underserved or vulnerable groups. In this regard, recently HUD restored Affirmatively Furthering Fair Housing rule that requires ``HUD to administer its programs and activities relating to housing and urban development in a manner that affirmatively furthers the purposes of the Fair Housing Act'', extending the existing non-discrimination mandates~\cite{2021RestoringCertifications}.

Our approach can also be extended to non-affirmative policies. This is possible by imposing constraints that ensure a topology has the same connections to all  protected groups within a score group. Such constraints are expressible as linear constraints and can be easily incorporated in Problem~\eqref{prob:MILP}. We demonstrate the result for fairness on race in Figure~\ref{fig:race-topo}. We observe that all individuals who belong to a certain queue, regardless of their race, are eligible for the same types of resources. However, as a result of combining the queues, the worst-case policy value across the racial groups decreases from 0.76 to 0.73 which still outperforms SQ and SQ (data) with worse-case value of 0.61 and 0.69, respectively. We defer the results for fairness by age to Appendix.

\section{Research Ethics and Social Impact}\label{sec:discussion}
Recently, there has been a significant growth in algorithms that assist decision-making across various domains~\cite{Rice2012DiscriminatoryColor, Goodman2018MachineReasoning, Simonite2020MeetCollege, Monahan2016RiskSentencing}. Homelessness is a pressing societal problem with complex fairness considerations which can benefit greatly from data-driven solutions. As empirical evidence on ethical side effects of algorithmic decision-making is growing, care needs to be taken to minimize the possibility of indirect or unintentional harms of such systems. 
We take steps towards this goal. Specifically, we propose \emph{interpretable} data-driven policies that make it easy for a decision-maker to identify and prevent potential harms. Further, we center our development around issues of \emph{fairness} that can creep into data from different sources such as past discriminatory practices. 
We provide a flexible framework to design policies that overcome such disparities while ensuring efficient allocations in terms of wait time and policy outcome.

There are also crucial consideration before applying our framework in real-world. Our approach relies on several key assumptions about the data. Specifically, the consistency assumption requires that there is only one version of PSH, RRH, and SO. In practice, different organizations may implement different variants of these interventions. For example, combining substance abuse intervention with PSH and RRH. Such granular information about the interventions, however, is not currently recorded in the data which may impact CATE estimates. Further, the exchangeability assumption requires that there are no unobserved confounders between treatment assignment and outcomes. Even though our dataset consists of a rich set of features for each individual, in practice, unobserved factors may influence the allocation of resources which calls for more rigorous inspection of service assignment processes. Unobserved confounders may lead to biased estimates of treatment effects which in turn impacts the allocation policies.
In addition, our dataset consists of samples from 16 communities across the U.S., which may not be representative of new communities or populations. Hence, the external validity of such policies should be carefully studied before applying to new populations. Finally, there are other domain-specific constraints that we have not considered as they require collecting additional data. For example, resources can not be moved between different CoCs. We leave such considerations to future work.

\begin{acks}
P. Vayanos and E. Rice gratefully acknowledge support from the Hilton C. Foundation, the Homeless Policy Research Institute, and the Home for Good foundation under the ``C.E.S. Triage Tool Research \& Refinement'' grant. P. Vayanos is also grateful for the support of the National Science Foundation under CAREER award number 2046230.
Finally, we thank the project's community advisory board for helpful discussions throughout the development of this work. 
\end{acks}





\bibliographystyle{ACM-Reference-Format}




\newpage
\appendix


\section{Appendix}\label{app:proof}

\subsection{Supplemental Material: Proof of Proposition~\ref{prop:policy_value}}

\begin{proof}
We let the $\bm X_q = 1$ be the event where $P(\bm X) = q$ ($\bm X_q = 0$ otherwise). Using this notation, we can write:
\begin{align*}
    V(\pi_M) & =  \EX\left[\sum_{r\in\mathcal R}\pi(r \mid \bm X)Y(r)\right] 
    \\
    & = \sum_{q \in \mathcal Q}\mathbb{P}(\bm X_q = 1)\EX\left[\sum_{r\in\mathcal R}\pi(r \mid \bm X)Y(r) \,\middle\vert\, \bm X_q = 1 \right] 
    \\
    & = \sum_{q \in \mathcal Q} \mathbb{P}(\bm X_q = 1)
    \EX\left[\sum_{r\in\mathcal R}\pi(r \mid \bm X)\left(Y(r) - Y(0)\right) \,\middle\vert\, \bm X_q = 1 \right] + 
    \mathbb{P}(\bm X_q = 1) \EX\left[\sum_{r \in \mathcal R}\pi(r \mid \bm X) Y(0) \,\middle\vert\, \bm X_q = 1\right]
    \\
    & = \sum_{q \in \mathcal Q} \mathbb{P}(\bm X_q = 1)
    \EX\left[\sum_{r\in\mathcal R}\pi(r \mid \bm X)\left(Y(r) - Y(0)\right) \,\middle\vert\, \bm X_q = 1 \right] + 
    \mathbb{P}(\bm X_q = 1) \EX\left[Y(0) \,\middle\vert\, \bm X_q = 1\right]
    \\
    & = \sum_{q \in \mathcal Q} \mathbb{P}(\bm X_q = 1)
    \EX\left[\sum_{r\in\mathcal R}\pi(r \mid \bm X)\left(Y(r) - Y(0)\right) \,\middle\vert\, \bm X_q = 1 \right] + C 
    \\
    & = \sum_{q \in \mathcal Q} \mathbb{P}(\bm X_q = 1)
    \sum_{r\in\mathcal R}\pi(r \mid \bm X)\EX\left[Y(r) - Y(0) \,\middle\vert\, \bm X_q = 1 \right] + C  
    \\
   & = \sum_{q \in \mathcal Q} \frac{\lambda_q}{\lambda_{\mathcal Q}} \sum_{r \in \mathcal R}\frac{f_{qr}}{\lambda_q}  \tau_{qr} + C
    \\
   & = \sum_{q \in \mathcal Q} \sum_{r \in \mathcal R} \frac{f_{qr}\tau_{qr}}{\lambda_{\mathcal Q}} + C, 
\end{align*}
where $C = \sum_{\q \in \sQ}\mathbb{P}(\bm X_q = 1) \EX\left[Y(0) \,\middle\vert\, \bm X_q = 1\right] =  \EX\left[Y(0)\right].$
\end{proof}

\subsection{Supplemental Material: Proof of Proposition~\ref{prop:feature-summarization}}
\begin{proof}
We first prove part one and show the conditional independence for each component $Y_r$ of the potential outcome vector. The proof is in the same vein as the balancing scores in the causal inference literature which is essentially a low-dimensional summary of the feature space that facilitates causal inference for observational data in settings with many features. For binary potential outcomes, we have
\begin{align*}
    \mathbb{P}(Y_{r} = 1\mid \bm S, \R) & =  \mathbb{E}[Y_{r} \mid \bm S, \R] \\
    & = \mathbb{E}\left[\mathbb{E}\left[Y_{\r} \mid \bm S, \R, \bm X \right]\,\vert\,\bm S, \R\right] \\
    & = \mathbb{E}\left[\mathbb{E}\left[Y_{r} \mid \bm S, \bm X \right]\,\vert\,\bm S, \R\right] \\
    & = \mathbb{E}\left[\mathbb{E}\left[Y_{r} \mid \bm X \right]\,\vert\,\bm S, \R\right] \\
    & = \mathbb{E}\left[S_r\,\vert\,\bm S, \R\right] \\
    & = S_r, 
\end{align*}
where the third line follows the assumption of the proposition and the fourth line holds since $\bm S$ is essentially a function of $\bm X$ and can be dropped. 
We also show
\begin{align*}
    \mathbb{P}(Y_{r} = 1\mid \bm{S}) & =  \mathbb{E}[Y_{r} \mid \bm S] \\
    & = \mathbb{E}\left[\mathbb{E}\left[Y_{r} \mid \bm S, \bm X \right]\,\vert\,\bm S\right] \\
    & = \mathbb{E}\left[\mathbb{E}\left[Y_{r} \mid  \bm X \right]\,\vert\,\bm S\right] \\
    & = \mathbb{E}\left[S_r\,\vert\,\bm S\right] \\
    & = S_r.
\end{align*}
We proved $\mathbb{P}(Y_{r} = 1\mid \bm S, \R) = \mathbb{P}(Y_{r} = 1\mid \bm S)$. 
We now prove the second part of the proposition. 
\begin{align*}
    \mathbb{P}\left( \mathbb{P}(R = r \mid \bm X = \bm x) > 0 \right) = 1 & \Rightarrow \mathbb{P}\left( \mathbb{P}(R = r, \bm X = \bm x) > 0 \right) = 1\\
\end{align*}
\begin{align*}
    \mathbb{P}\left( \mathbb{P}(R = r, \bm X = \bm x) > 0 \right) & =  \mathbb{P}\left( \mathbb{P}(R = r, \bm X = \bm x, \bm S = \bm s) > 0 \right) \\
    & \leq \mathbb{P}\left( \mathbb{P}(R = r, \bm S = \bm s) > 0 \right). 
\end{align*}
It follows that $\mathbb{P}\left( \mathbb{P}(R = r, \bm S = \bm s) > 0 \right) = 1$ for all values of $\bm s$. 
\end{proof}

\section{Supplemental Material: Computational Results}\label{app:results}


\noindent\textbf{HMIS Data Preparation. }
We used HMIS dataset collected between 2015 and 2017 across 16 communities in the United States. The dataset contains 10,922 homeless youth and 3464 PSH and RRH resources combined. 
We removed all those with veteran status (54 data points), pending and unknown outcomes (4713 data points). We grouped Hawaiian/Pacific Islander, Native American, Hispanics, Asian under `Other' category as no significant statistical inference can be made on small set of observations within each individual category. Further, we removed 6 data points with no gender information. We use a median date 08/13/2015 as the cut-off date to separate train and test sets.

\begin{figure}[b!]
    \centering
    \includegraphics[width = 0.5 \textwidth]{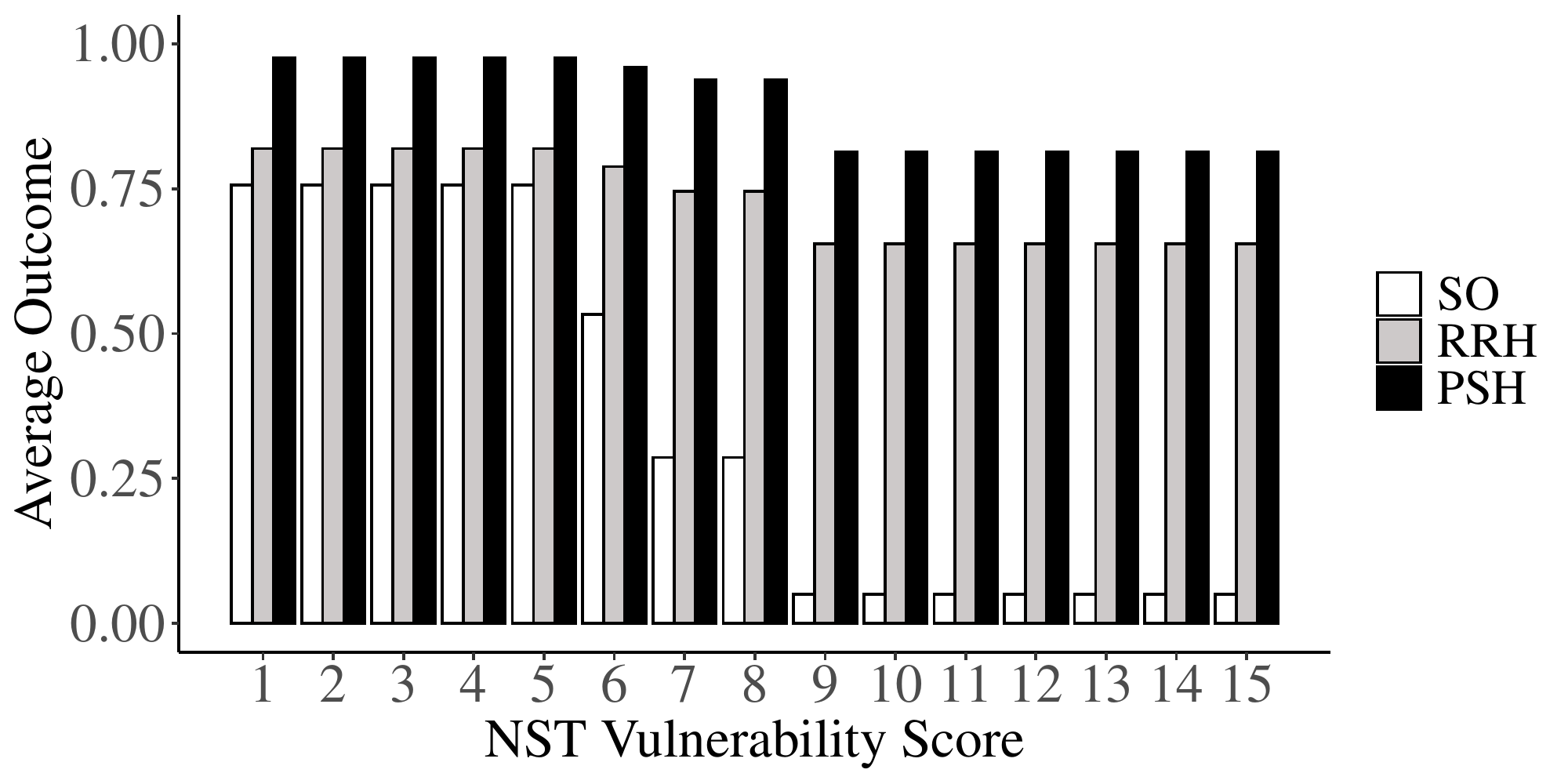}
    \caption{Probability of exiting homelessness across the NST score range estimated using the DR method.}
    \label{fig:outcome-finalscore}
\end{figure}

\noindent\textbf{Outcome Estimation. }
Figure~\ref{fig:outcome-finalscore} depicts the average outcome across different score values $\mathbb E[Y(\r) \mid S = s] \; \forall \r \in \sR,$ using the DR estimate. Under SO, after $S = 8$, there is a significant drop in average outcome. Average outcomes under PSH and RRH also exhibit a decline with score. However, they remain highly effective even for high-scoring youth. 

\begin{figure}[t!]
    \begin{subfigure}[b]{0.95\textwidth}
    \centering
         \includegraphics[width=\textwidth]{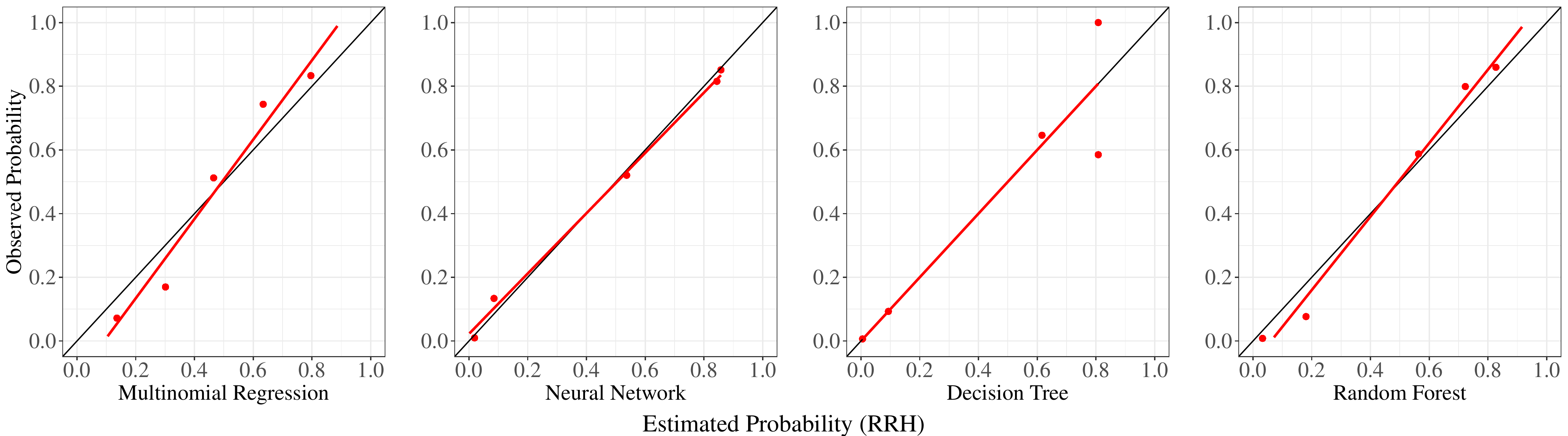}
    \end{subfigure}
    \begin{subfigure}[b]{0.95\textwidth}
    \centering
         \includegraphics[width=\textwidth]{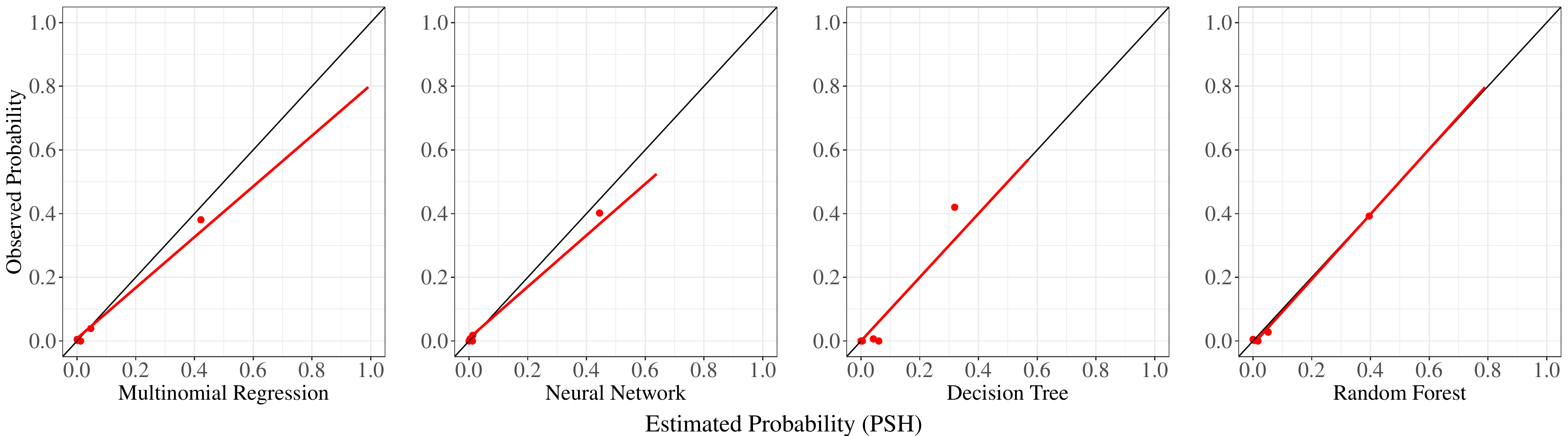}
    \end{subfigure} 
    \caption{Reliability diagram of propensity estimation, RRH (top) and PSH (bottom). }
    \label{fig:reliability-propensity}
\end{figure}

\noindent\textbf{Propensity Score. }
In order to evaluate different policies using IPW and DR methods, we estimated the propensity scores, i.e., $\pi_0(\R = \r \mid \vX = \vx)$. Table~\ref{tab:propensity} summarizes the accuracy across different models. We consider two models, one that uses only the NST score and one that uses the entire set of features in the data. We observe that, even though the policy recommendations only use NST score, including other features help improve the accuracy. In addition, the decision tree and random forest are the top-performing models. Although random forest exhibits over-fitting (in-sample accuracy = 99.6\%) its out-of-sample accuracy (79.3\%) outperforms other models. In addition to accuracy, the propensity models should be well-calibrated. That is, the observed probability should match the predicted probability. We plot the reliability diagrams in Figure~\ref{fig:reliability-propensity}, where $y-$axis is the observed probability in the data and the $x-$axis is the predicted value. The dots correspond to values of different bins. A well-calibrated model should lie on the $y = x$ diagonal line. 

\begin{table}[t!]
    \centering
\begin{tabular}{P{2cm}P{4.0cm}P{3cm}P{4.0cm}}
\toprule
& Model & In-Sample Accuracy (\%) & Out-of-Sample Accuracy (\%) \\
\hline
\multirow{4}{*}{NST Score} & Multinomial Regression & 72.5 & 73.7 \\
\cmidrule(lr){2-4}
& Neural Network & 76.4 & 76.5 \\
\cmidrule(lr){2-4}
& Decision Tree & 76.3 & 76.2 \\
\cmidrule(lr){2-4}
& Random Forest & 76.4 & 76.3 \\ 
\bottomrule
\multirow{4}{*}{All Features}  & Multinomial Regression & 75.4 & 73.5 \\
\cmidrule(lr){2-4}
& Neural Network & 80.4 & 77.2 \\
\cmidrule(lr){2-4}
& Decision Tree & 79.2 & 78.5 \\
\cmidrule(lr){2-4}
& Random Forest & 99.7 & 79.3 \\ 
\bottomrule
\end{tabular}
\caption{Prediction accuracy for propensity estimation using HMIS data.}
\label{tab:propensity}
\end{table}
As seen in Figure~\ref{fig:reliability-propensity}, random forest and neural network models have relatively better calibration property. Finally, in our model selection, we take fairness considerations into account. In particular, we study the calibration of the models across different demographic groups for which fair treatment is important. Since ultimately we use the probability estimates, not the binary prediction, it is important to ensure that across different demographic groups, the models are well-calibrated. We adopted test-fairness notion~\cite{Chouldechova2017FairInstruments}. We fit a model to predict the resource one receives, based on the predicted propensities and demographic features. In a well-calibrated model across demographic groups, the coefficients of the demographic attributes should not be statistically significant in the prediction. For the predicted values of the random forest model none of the demographic attributes coefficients were found to be statistically significant. In addition, the model were calibrated within groups with coefficient near 1. Regression results are summarized in Table~\ref{tab:calibration-random-forest}. Hence, we chose random forest as the model of historical policy $\pi_0$. 
\begin{table}
    \centering
\begin{tabular}{cc}
    \begin{tabular}{lllll}
    \toprule
    Coeffs. & Estimates & $p$-value \\
    \hline
    Intercept & -0.012 & 0.066 \\
    PSH pred. & \textbf{0.985} & <2e-16 \\
    Race = 2 & 0.011 & 0.204 \\
    Race = 3 & 0.002 & 0.803 \\
    Gender = 2 & -0.006 & 0.469 \\
    Age = 2 & 0.006 & 0.485 \\ 
    \bottomrule
    \end{tabular}
    &
    \begin{tabular}{lllll}
    \toprule
    Coeffs. & Estimates & $p$-value \\
    \hline
    Intercept & -0.054 & 5.7e-05 \\
    RRH pred. & \textbf{1.125} & < 2e-16 \\
    Race = 2 & -0.007 & 0.586 \\
    Race = 3 & -0.014 & 0.394  \\
    Gender = 2 & 0.000 & 0.987 \\
    Age = 2 & -0.003 & 0.813  \\ 
    \bottomrule
    \end{tabular}
\end{tabular}
    \caption{Propensity calibration within group for PSH (left) and RRH (right) of random forest model. None of the coefficients of the demographic attributes are found to be significant. In addition, the coefficient associated with the predicted probability is close to 1 in both models, suggesting that the model is well-calibrated even when we control for the demographic attributes.}
    \label{tab:calibration-random-forest}
\end{table}

\noindent\textbf{Outcome Estimation. }In the direct method, one estimates the (counterfactual) outcomes under different resources by fitting the regression models $\mathbb P(Y \mid \vX = \vx, \R = \r) \; \, \forall \r \in \sR.$ For model selection, we followed the same procedure as propensity score estimation. Table~\ref{tab:outcome-est-reward-complex} summarizes the accuracy of different models for each type of resource. 
\begin{table}[ht]
    \centering
\begin{tabular}{P{3.5cm}P{3.5cm}P{1.0cm}P{1.0cm}P{1.0cm}}
\toprule
& Model & PSH & RRH & SO \\
\hline
\multirow{4}{*}{NST} & Logistic Regression & 83.1 & 78.8 & 90.0 \\
\cmidrule(lr){2-5}
& Neural Network & 83.9 & 78.9 &  90.0 \\
\cmidrule(lr){2-5}
& Decision Tree & 83.9 & 78.9 & 90.0 \\
\cmidrule(lr){2-5}
& Random Forest  & 83.1 & 78.6 & 90.0 \\ 
\bottomrule
\multirow{4}{*}{NST + Demographic} & Logistic Regression & 83.1 & 78.8 &  90.0 \\
\cmidrule(lr){2-5}
& Neural Network & 81.6 & 78.3 & 90.3 \\
\cmidrule(lr){2-5}
& Decision Tree & 83.9 & 78.8 & 90.0\\
\cmidrule(lr){2-5}
& Random Forest  & 83.9 & 78.1 & 90.0\\ 
\bottomrule
\multirow{4}{*}{All Features} & Logistic Regression & 81.9 & 82.2 & 90.3  \\
\cmidrule(lr){2-5}
& Neural Network & 83.9 & 78.8 & 86.8  \\
\cmidrule(lr){2-5}
& Decision Tree & 74.3 & 81.1 & 90.0 \\
\cmidrule(lr){2-5}
& Random Forest  & 83.9 & 81.4 & 90.0 \\ 
\bottomrule
\end{tabular}
\caption{Out-of-Sample Accuracy (\%) of different outcome estimation models (outcome definition in Figure~\ref{fig:success-definition}).}
\label{tab:outcome-est-reward-complex}

\end{table}

\begin{figure}[t!]
    \begin{subfigure}[b]{0.95\textwidth}
    \centering
         \includegraphics[width=\textwidth]{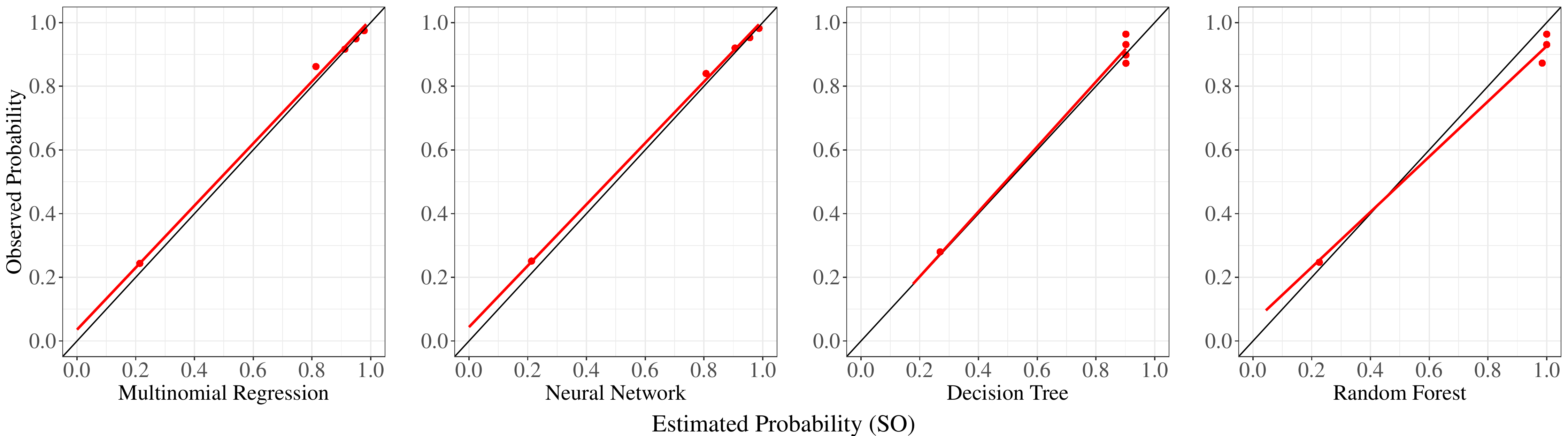}
    \end{subfigure}
    \begin{subfigure}[b]{0.95\textwidth}
    \centering
         \includegraphics[width=\textwidth]{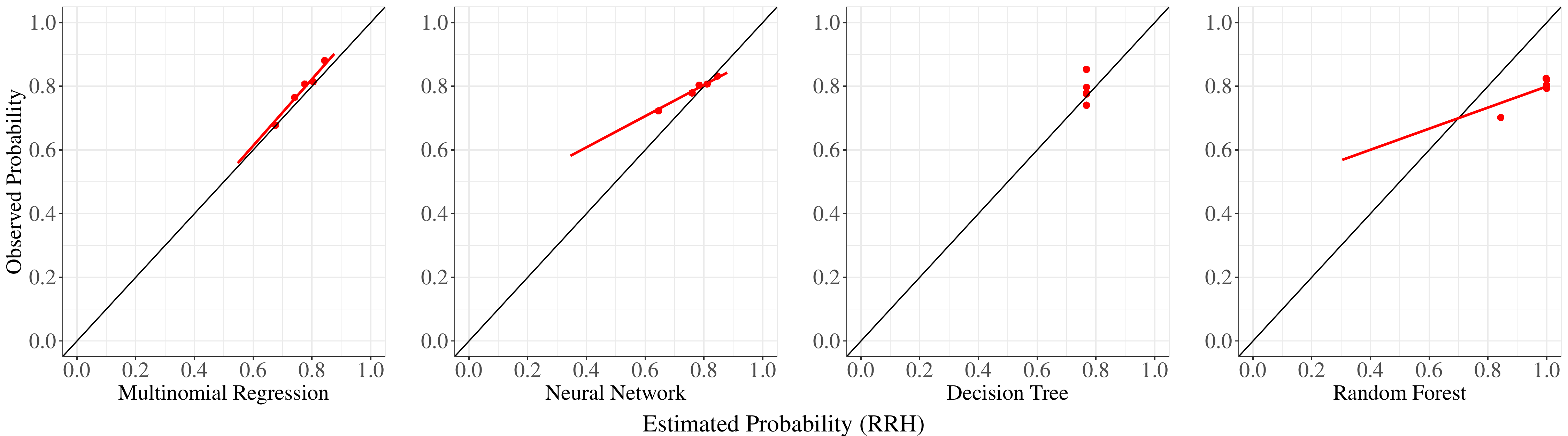}
    \end{subfigure}
    \begin{subfigure}[b]{0.95\textwidth}
    \centering
         \includegraphics[width=\textwidth]{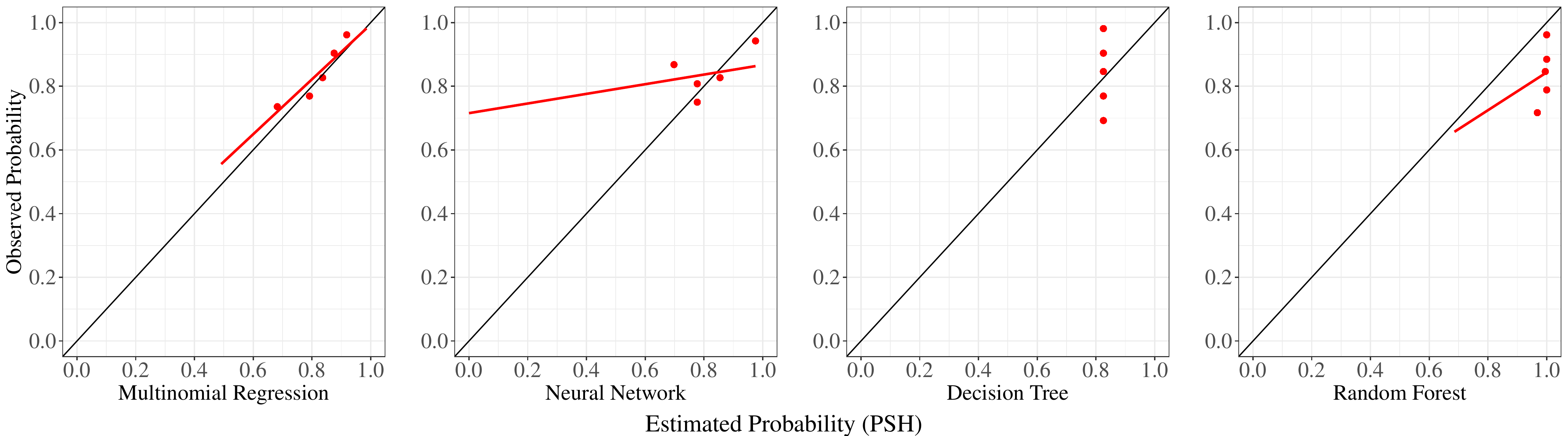}
    \end{subfigure} 
    \caption{Reliability diagram of outcome estimation, RRH (top) and PSH (bottom). }
    \label{fig:reliability-outcome}
\end{figure}
\begin{table}[t]
    \centering
\begin{tabular}{ccc}
    \begin{tabular}{lllll}
    \toprule
    Coeffs. & Estimates & $p$-value \\
    \hline
    Intercept & 0.147 & 0.489 \\
    PSH pred. & \textbf{0.853} & 0.000 \\
    Race = 2 & -0.021 & 0.666 \\
    Race = 3 & -0.061 & 0.324 \\
    Gender = 2 & 0.003 & 0.954 \\
    Age = 2 & 0.079 & 0.202 \\ 
    \bottomrule
    \end{tabular}
    \begin{tabular}{lllll}
    \toprule
    Coeffs. & Estimates & $p$-value \\
    \hline
    Intercept & -0.122 & 0.645 \\
    RRH pred. & \textbf{1.172} & 0.000 \\
    Race = 2 & 0.028 & 0.386 \\
    Race = 3 & 0.025 & 0.504 \\
    Gender = 2 & -0.021 & 0.433 \\
    Age = 2 & 0.003 & 0.931 \\ 
    \bottomrule
    \end{tabular}
    &
    \begin{tabular}{lllll}
    \toprule
    Coeffs. & Estimates & $p$-value \\
    \hline
    Intercept & 0.035 & 0.148  \\
    SO pred. & \textbf{0.974} & <2e-16  \\
    Race = 2 & -0.000 & 0.973  \\
    Race = 3 & 0.023 &  0.226  \\
    Gender = 2 & -0.008 & 0.618 \\
    Age = 2 & -0.011 & 0.542 \\ 
    \bottomrule
    \end{tabular}
\end{tabular}
    \caption{Outcome calibration within group for PSH (left) and RRH (right) of logistic regression model. None of the coefficients of the demographic attributes are found to be significant. In addition, the coefficient associated with the predicted probability is close to 1 in both models, suggesting that the model is well-calibrated even when we control for the demographic attributes.}
    \label{tab:calibration-logistic}
\end{table}
Considering the reliability diagrams in Figure~\ref{fig:reliability-outcome}, we observe that logistic regression models are well-calibrated across different resources. We also investigated test-fairness of logistic regression where we fit the observed outcome against the predicted outcome and demographic features. Results are summarized in Table~\ref{tab:calibration-logistic}. As seen, the coefficients of demographic features are not significant, suggesting that test-fairness is satisfied.

\noindent\textbf{Optimal Matching Topology for Fairness over Age.} 

\begin{figure}[t!]
    \begin{subfigure}[t]{0.33\textwidth}
    \centering
         \includegraphics[width=\textwidth]{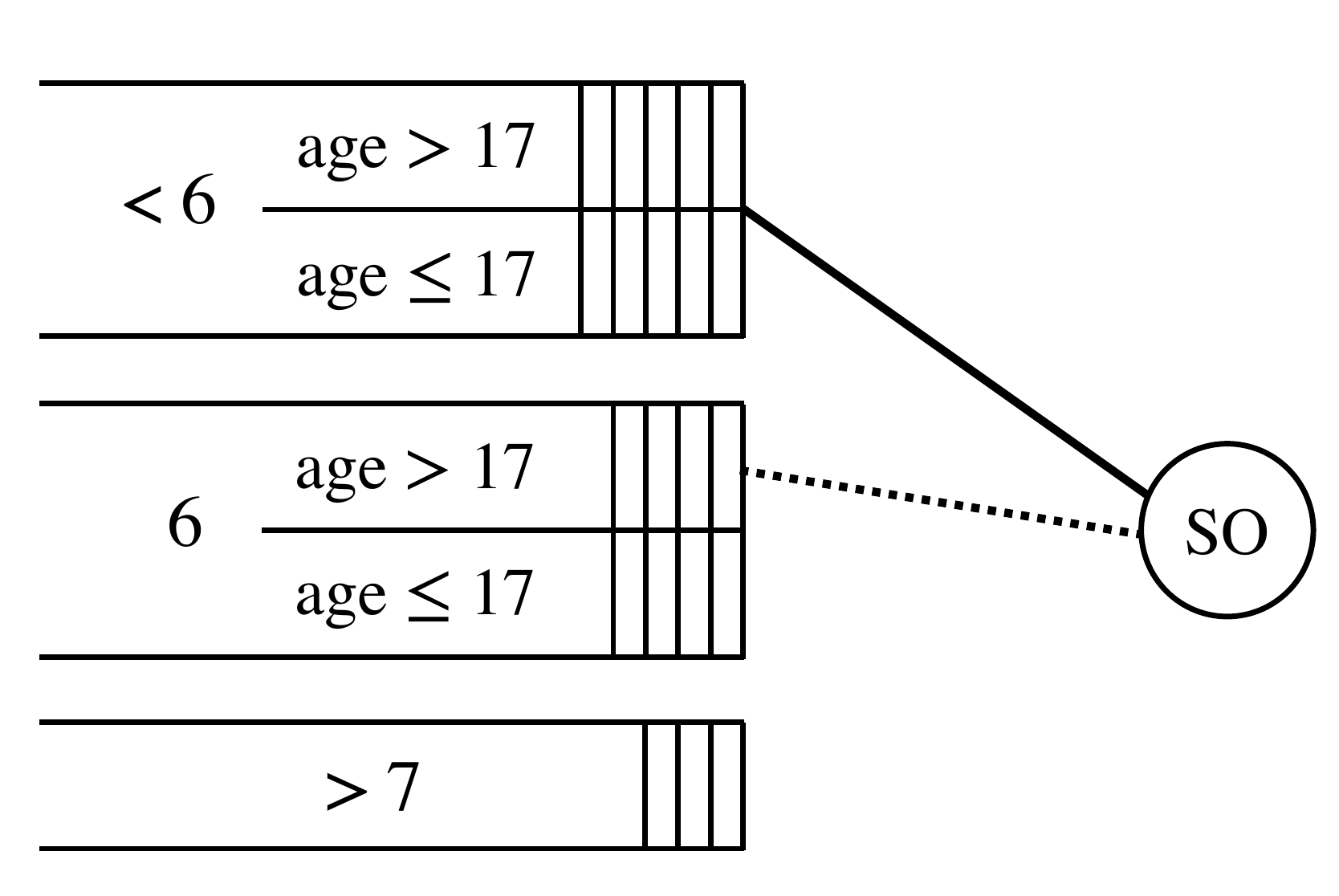}
         \label{fig:opt-age-SO}
    \end{subfigure}\hfill
    \begin{subfigure}[t]{0.33\textwidth}
    \centering
         \includegraphics[width=\textwidth]{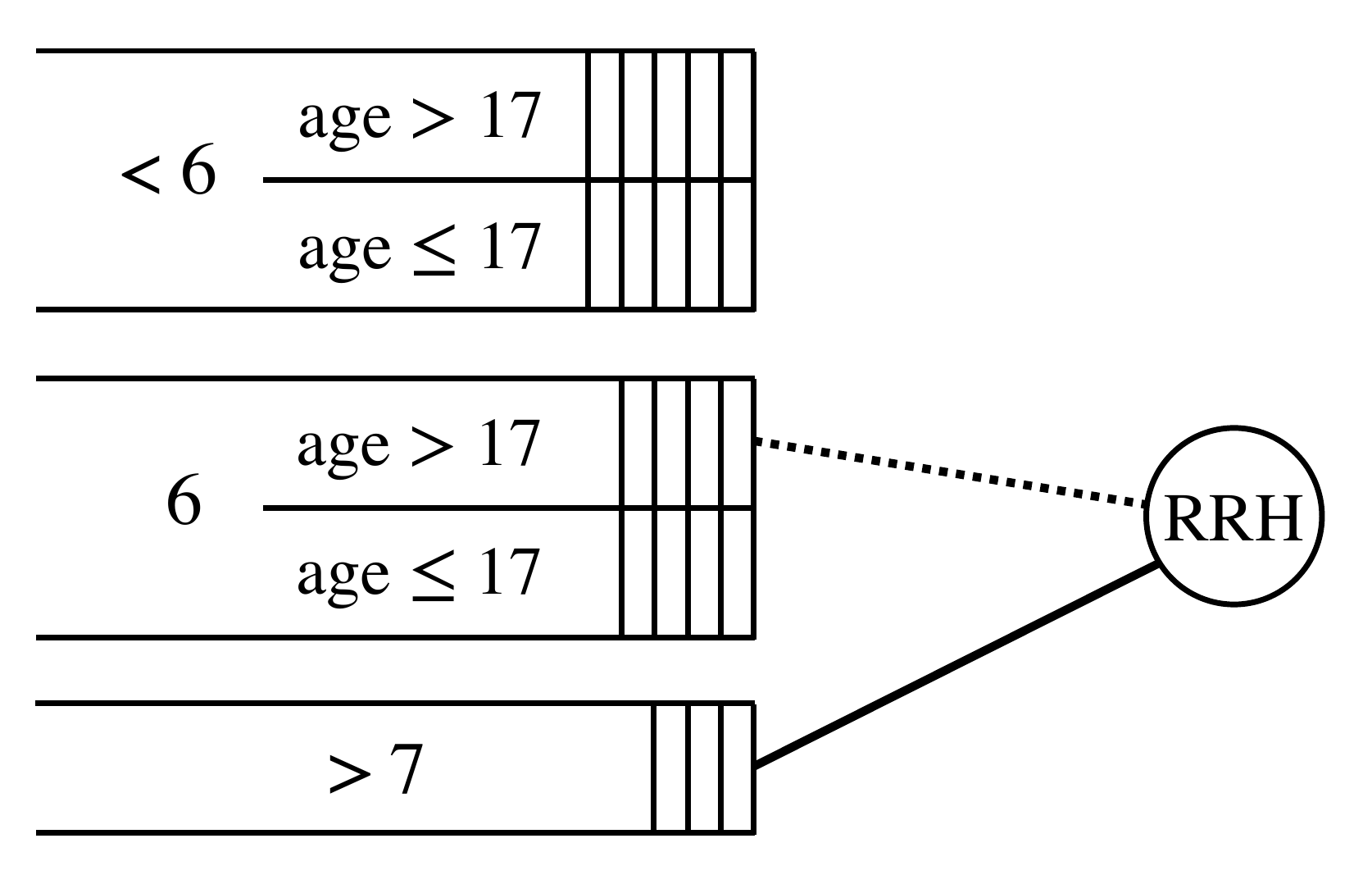}
         \label{fig:opt-age-RRH}
    \end{subfigure}\hfill  
    \begin{subfigure}[t]{0.33\textwidth}
    \centering
         \includegraphics[width=\textwidth]{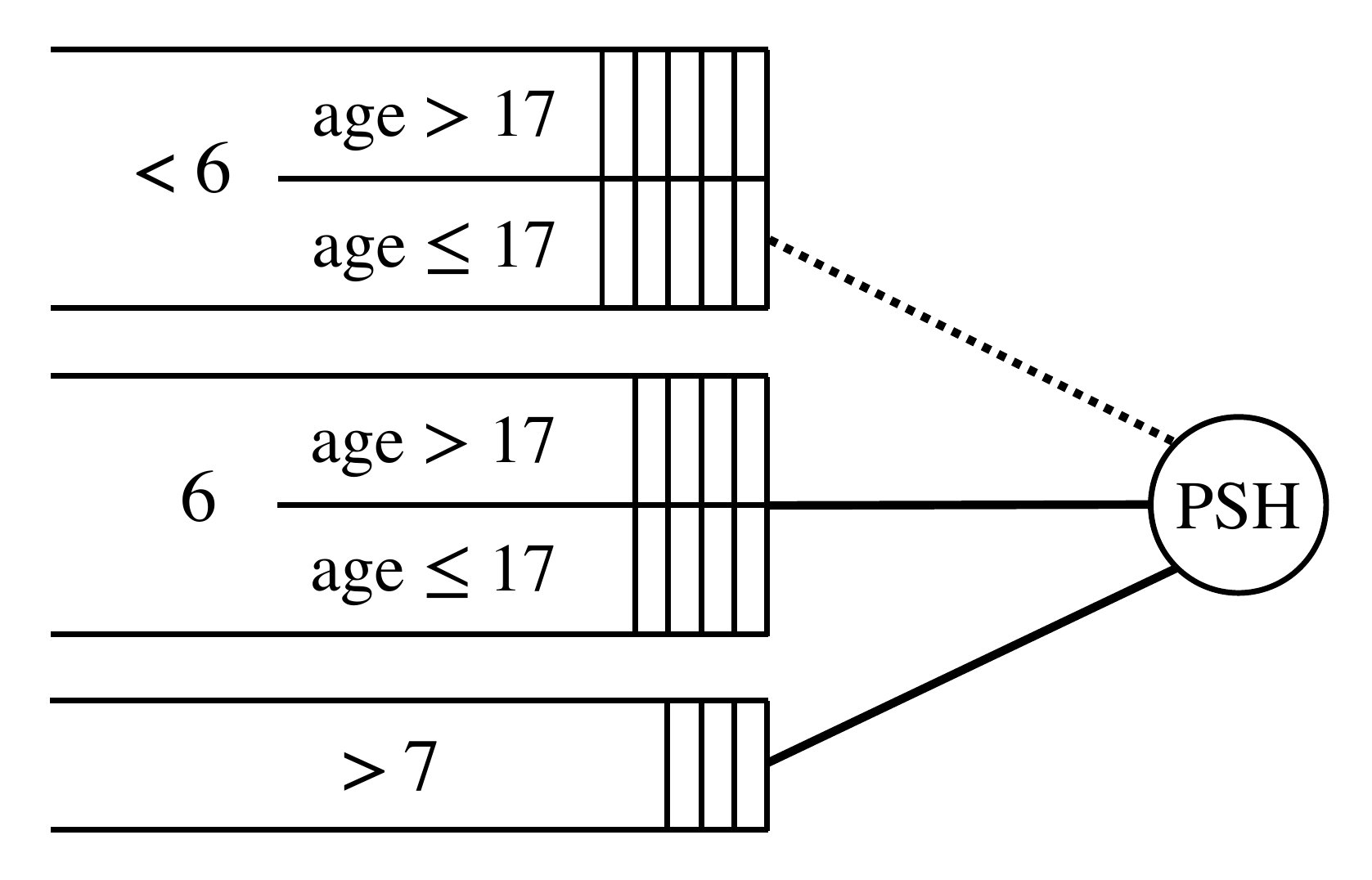}
         \label{fig:opt-age-PSH}
    \end{subfigure} 
    \caption{The matching topology split by resource type: left (SO), middle (RRH) and right (PSH). The solid line indicates that the resource is connected to the entire queue. The dotted line indicates connection to a sub-group within the queue. For example, in the left figure, SO is only connected to the individuals with NST = 6 and age over 17.}
    \label{fig:opt-age}
\end{figure}

\begin{figure}[t]
    \centering
    \includegraphics[width=0.3\textwidth]{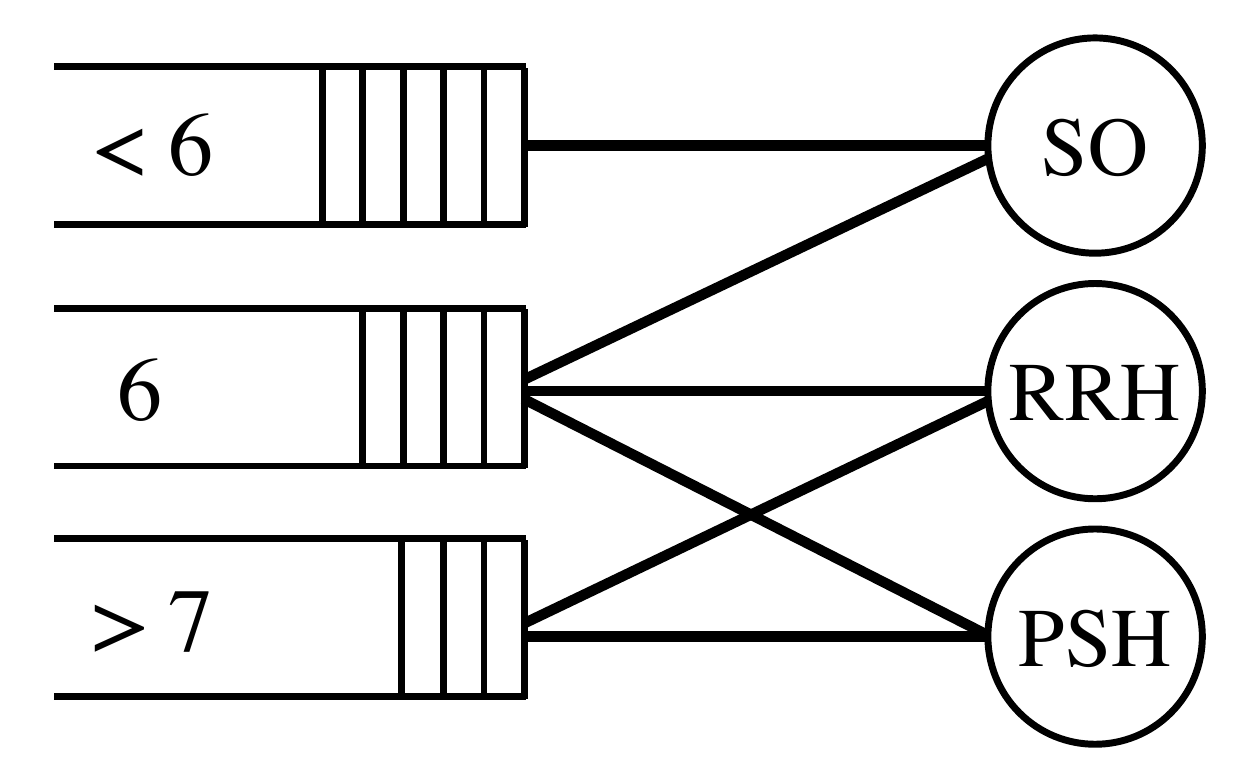}
    \caption{Fair topology (age)}
    \label{fig:opt-age-noDependece}
\end{figure}

Figure~\ref{fig:opt-age} depicts the policies when fairness over age is imposed. According to this figure, across all score values youth below 17 years are eligible for PSH. On the other hand, mid- and high-scoring youth over 17 years old, are eligible for PSH. 
We further imposed constraints to ensure within each score group, the connections are the same for different age groups. Figure~\ref{fig:opt-age-noDependece} illustrates the resulting matching topology, according to which individuals who score above 7 are eligible for RRH and PSH, regardless of their age. Those who score 6 are eligible for all three resource types. Finally, All youth with score below 6 are only eligible for SO. We observe that all individuals who belong to a certain queue, regardless of their age, are eligible for the same types of resources. As a result of combining the queues that depended on age, the worst-case policy value across the age groups decreased from 0.74 to 0.69 which still outperforms the SQ (data) with worst-case performance of 0.64. 


\end{document}